\documentclass[pdflatex,smallcondensed,Numbered]{sn-jnl}

\usepackage{graphicx}
\usepackage{multirow}
\usepackage{amsmath,amssymb,amsfonts}
\usepackage{amsthm}
\usepackage[title]{appendix}
\usepackage{xcolor}
\usepackage{textcomp}
\usepackage{booktabs}
\usepackage{tikz}
\usepackage{algorithm}
\usepackage{algpseudocode}
\usetikzlibrary{arrows.meta,positioning}
\usepackage{pifont}
\usepackage{manyfoot}
\usepackage{url}

\newtheorem{proposition}{Proposition}
\newtheorem{corollary}{Corollary}

\title{Contract-Explicit Evaluation: From Auditing Learned Readers to Scoped Certificates}

\author[1]{\fnm{Hao} \sur{Lu}}\email{luhao\_0210@163.com}

\author*[1]{\fnm{Weixing} \sur{Zhu}}\email{zwx@aeu.edu.cn}

\author[1]{\fnm{Ruiyang} \sur{Li}}\email{lry130923@aeu.edu.cn}

\affil*[1]{Army Engineering University of PLA, Nanjing, Jiangsu, China}

\abstract{Evaluation readings of document-level relation extraction
(DocRE) systems depend jointly on the model and the \emph{evaluation
contract}: gold version, scoring function, threshold. Rankings
reverse across contracts, yet no existing method determines which
comparisons survive the reversal. We address this gap with
contract-explicit evaluation: a preregistered five-step protocol
(Declare, Measure, Scope, Certify, Hedge), operationalizing the
classical concordance decomposition of Kendall's
rank correlation, that scopes certifiable
regions and delivers distribution-free census certificates across
two ecosystems. In the DocRE audit (44 training runs of one
in-house architecture family), the scoring dimension scopes to a
census certificate (three flagged instances, Clopper--Pearson upper
bound $0.146$); the confirmatory record includes registered
negatives, and the audit prices its own simplification finding---the
calibration dial's entire margin is one withdrawn commit. In an
LLM-judge census (17 judges, $49{,}938$ battles), $12.25\%$ of
leaderboard comparisons require hedging at the pre-declared
split criterion (bound $0.124$; panel-robust $0.143$), with an
exploratory two-tier reading certifying $69\%$ at a bound
$\le 0.072$ and a validity decomposition against released human
labels ($65\%$ of contested battles are judge error, not
ambiguity). The cross-model certificate is withdrawn on real
confidence scores; the failure spectrum maps the boundary. The
protocol, map, ledger, and per-instance artifacts are released.}


\begin{document}

\maketitle

\section{Introduction}
\label{sec:intro}

This paper poses contract-explicit evaluation: performance readings
of intelligent systems are joint properties of the models and
\emph{evaluation contracts}. The paper delivers both an audit of one
ecosystem (Parts~1--2) and the method that generalizes it (Part~3).
The audit's instrument is a \emph{learned evaluation reader}, a
per-document weighting network over multi-contract readings of a
frozen prediction pool, trained toward the contract consensus and
wrapped in a risk-controlled decision layer. The method movement
distills the audit's discipline into five steps (Declare, Measure,
Scope, Certify, Hedge) anchored by a scoping identity that states
where certificates may hold. The learning problem exists because the
readings being aggregated are not properties of the model alone.

Performance readings
in document-level relation extraction (DocRE) are
routinely cited as evidence of capability: the intra--inter sentence gap of a
model, its micro-F1 under a chosen gold standard, its ranking against a
baseline. The practice has a precise institutional shape. Leaderboards and
ablation tables report a single contracted number per
system~\cite{yao-etal-2019-docred}; the
intra--inter gap in particular has served as a structural capability claim
since the earliest graph- and segment-based DocRE
architectures~\cite{yao-etal-2019-docred,chen-etal-2021-docunet}, and an evidence-aware line of
work exists precisely to repair the cross-sentence weakness that signal
flags~\cite{xie-etal-2022-eider}; the
community's evaluation habits inherit the convention: a model is said to ``reason
across sentences'' when its inter-sentential reading improves under one
declared gold, scorer, and threshold. What these habits presuppose---that
the contracted number is a property of the model alone---is exactly what
contract-sensitivity studies refute. Companion work~\cite{lu2026evaluation} established that such
readings are \emph{joint properties of models and evaluation contracts}---the
gold-standard version $g$, the scoring function $s$, and the thresholding rule
$\tau$ under which frozen predictions are read. With model predictions held
fixed, exchanging one legal gold standard for another reverses method
orderings---in a seven-family pool that includes companion retrains,
and independently for LLM-based rankings under the Re-DocRED
rescore~\cite{sun2026consistre}. The frozen registry of the companion submission records 13
rank flips among 57 checked comparisons, 11 of whose flip-difference confidence
intervals exclude zero---and swapping judges or annotator pools
produces the same instability on other
axes~\cite{han2026swapping,jha2026whosegold}.

The companion work's remedy is declarative: an audited evaluation reading
(AER) that marginalizes over a declared contract set with equal weights,
declared ties, and a seed prescription. But two of its own findings delimit
what fixed procedures can achieve. First, \emph{member selection is itself a
contract choice}: leave-one-member-out moves the AER reading across intervals,
and a two-member subset outranks the registered three-member
combination~\cite{lu2026evaluation}. Second, the union-of-ties rule attains
89\% flip coverage only by accepting a 72--91\% false-tie rate---a
coverage--precision trade-off that the declarative procedure accepts but does
not optimize. Fixed aggregation, in other words, is a point on a trade-off
curve chosen by convention rather than by learning.

This paper treats that trade-off as a learning problem. We freeze the
prediction pool (44 runs from 7 generator families on the 498-document
double-gold DocRE dev split---the DocRED dev corpus co-aligned with the Re-DocRED dev revision.
Precisely: Re-DocRED's dev set is the 998-document DocRED dev split
with corrected annotations; its 500-document evaluation portion is
sampled from those documents, and exactly two of them lack a
same-title DocRED counterpart, leaving 498 documents on which all
three gold revisions ($g_D$, $g_R$, $g_{R'}$) are computable (the
two dropped document IDs are released in the supplement; on length
and gold-relation count the 498 are indistinguishable from the full
dev set, median 8 sentences and 11 relations in both)---). On this pool we learn a
\emph{contract-marginalized
aggregator}: per-document features (cross-generator disagreement, relation
frequencies, document length, near-threshold density, contract reading
vectors, and margin/arm proxies derived from the companion's leverage theory)
condition a weight network that mixes the readings of the available contracts
for each generator. Training optimizes a \emph{contract-consensus} objective
---the signed median of contract-level paired differences, with margins
softened by a tie-aware kernel in the spirit of tie-aware ranking
objectives~\cite{he2018hashing}---plus a leave-one-contract-out flip penalty,
under an explicit held-out contract switch: the aggregator trains on gold-$g_D$
side contracts and is evaluated on the gold-$g_R$ side it never saw.

\noindent\textbf{Contributions.}
\begin{enumerate}
\item \textbf{Audit protocol.} A preregistered, five-step protocol for
\emph{learning evaluation readers}---(a) an ex-ante exchangeability check
(do the contracts read one quantity?), (b) registry power sizing (is the
flip count certifiable at $\alpha{=}0.05$?), (c) a frozen readout table
(criteria before the run), (d) a confirmatory run reported verbatim
(registered negatives included), and (e) post-hoc robustness probes
(decode calibration, consensus stability, decision gates). Each step
prevents a documented failure: unexamined exchangeability, test shopping,
false certainty, artifact flips, and coverage deficits
(Sections~\ref{sec:formulation}--\ref{sec:discussion}).
\item \textbf{Problem and instance.} We formalize robust evaluation
reading as a dual objective---ranking stability under contract switch and
reading quality---over multi-contract readings of a frozen prediction
pool, and instantiate it as a document-conditioned, generator-aware
weighting network trained with a consensus-signed soft-tie loss and a
leave-one-contract-out stability penalty, in which the fixed baselines
(mean-of-$z$ marginalization, median, best single contract,
union-of-ties) appear as degenerate points. A fidelity trade-off
proposition characterizes what the consensus target can and cannot
deliver (Sections~\ref{sec:formulation}--\ref{sec:method}).
\item \textbf{Evaluation suite.} Bitwise reproduction of the companion's
readings as a loading gate; registry coverage and false-tie under the
frozen tie-declaration rule; \emph{hard-ranking fidelity}; and a
coverage--false-tie frontier over confidence levels
(Section~\ref{sec:protocol}).
\item \textbf{Positioning.} To our knowledge, no prior work trains a
reader over evaluation-contract readings of a frozen prediction pool with
acceptance criteria frozen before the confirmatory run; what the audit
certifies, on its own evidence, is the protocol and the readout
discipline, not the necessity of the learned components---the
calibration dial's entire margin on this pool is one withdrawn
correct commit (Section~\ref{sec:discussion}), a simplification finding
we report rather than hedge.
\item \textbf{Findings.} The instance holds the lowest false-tie count
among hedging readers, one pair short of the preregistered coverage
floor---$47$--$48$ vs.\ $65$
for the best fixed baseline on the confirmatory registry---and consensus
fidelity perfect on every pair it ranks---a
training-sanity property---replicated at preregistered seeds on
the fourteen-family, 52-flip registry v3 ($43/43$; held-out $g_R$
agreement $26/43$, as the trade-off proposition requires). A post-hoc
Fisher contrast on the readers' self-selected, unequal hard-ranked sets
($43/43$ vs.\ $8/24$ for the $g_D$-native reader; $p{=}6.5{\times}10^{-10}$)
is reported for completeness; its like-for-like control is the
selection-matched analysis (the learned reader's $31$ hard-ranked pairs
form a strict superset of the native $15$, all consensus-consistent;
Section~\ref{sec:confirmv2}). The aggregate is
consensus-leaning by construction and does not track the held-out
$g_R$ native ordering---a structural property, not a defect. A preregistered
conformal layer returns a registered negative on the unconditional
cross-reference guarantee: contested pairs are precisely where audit
references fail to be exchangeable. A preregistered gate layer that
hedges exactly those conflicts passes the guarantee's
cross-reference transfer check at the realized rate---which a
reference-copying no-op nearly matches---at a loosened second-cycle
$\alpha^{*}{=}0.20$ (realized rates $0.10$/$0.00$; failing the
originally frozen $0.10$); the margin over that no-op is
non-inferior coverage, with a false-tie cost
that now also sits below the best fixed baseline---the choice of
audit reference is itself a first-order decision. Transfer to
TACRED~\cite{zhang-etal-2017-tacred} (a relation-extraction benchmark
with conflicting gold revisions) exposes a
clean applicability boundary: consensus aggregation requires
exchangeable readings, not conflicting gold revisions
(Section~\ref{sec:experiments}).
\item \textbf{Method and certificates (Part~3).} The audit
generalized into a five-step protocol---Declare, Measure, Scope,
Certify, Hedge---grounded in the classical concordance decomposition
of Kendall's $\tau$ (flip density equals $(1-\tau)/2$) operationalized
as a scoping principle, a scoping map with
quorum arithmetic, a delivered census certificate on $44$ training runs
of one in-house architecture family (three flagged instances, Clopper--Pearson (CP)
upper bound $0.146$), and a
failure-spectrum calibration of the boundary
(Section~\ref{sec:method5}).
\item \textbf{Artifacts.} Loader, aggregator, evaluation harness,
per-document readings and weights, all bitwise-anchored to released
frozen counts (Section~\ref{sec:protocol}); preregistrations
v1--v7 with amendments (v5: conventions protocol and the per-cell
Bonferroni bound; v6: pool growth to $n{=}44$ with the split-sample
prospective variant; v7: the LLM-judge instantiation of this
section), the certifiability map, and per-instance
reading tables (Section~\ref{sec:method5}).
\end{enumerate}

\paragraph{Why the title runs from audit to certificates.}
The direction is the paper's argument, not a narrative device.
Auditing the learned reader is how the certificate method was
refined. The registered negatives are what shaped the protocol's
failure modes---each hedge the audit needed became a protocol step
(Declare, Measure, Scope, Certify, Hedge); the measured contract
hierarchy located the certifiable region before any certificate was
sought; and the audit's simplification finding separated the
instrument from the contribution, leaving the discipline itself as
the deliverable. Auditing learned readers is how scoped certificates
were found---and where they hold is what the certificate map now
states.

The rest of this paper is structured as follows.
Section~\ref{sec:related} reviews the evaluation-contract and
certification lines of work. Sections~\ref{sec:formulation}
through~\ref{sec:discussion} form Part~1: the problem formulation, the
learned reader, the evaluation protocol, the experimental record, and
the discussion of its limits. Section~\ref{sec:part2} is Part~2: the
registry extension and the preregistered conformal decision layer,
with its registered negatives and the gate layer that repairs them.
Section~\ref{sec:method5} is Part~3: the five-step method, the scoping
identity, the certifiability map, the worked certificate, and the
failure spectrum. Section~\ref{sec:conclusion} concludes and outlines
future work.

\section{Related work}
\label{sec:related}

\paragraph{Judge-side learned aggregation.}
Closest in method shape, recent work learns aggregation weights over
\emph{judge outputs}. VoirDire derives a single global linear aggregation from
unlabeled intervention pairs on judge scores~\cite{shin2026voirdire}; AGORA
learns a global per-judge trust weight via item-response
theory~\cite{ahn2026agora}; am-ELO proves order-invariance of a Bradley--Terry
MLE over pairwise battles~\cite{liu2025amelo}; and acyclic preference
evaluation ensembles multiple evaluators preference graphs to excise
cycles~\cite{hu2024acyclic}. All four learn or derive aggregation weights over judge outputs, but they
differ in what varies: VoirDire's intervention pairs are perturbations of
surface attributes (an axis structurally parallel to our contract
switch), AGORA's item-response model carries per-item latent parameters
(not a single global weight), and am-ELO establishes order-invariance of
a Bradley--Terry maximum-likelihood estimate under sample reordering.
None of them operates on
thresholded structural decisions under gold-standard exchange, and none
is evaluated on ranking stability across a held-out contract. Our
corpus-learned variant (Section~\ref{sec:ablations}; it collapses
to $sd{=}0$) is a direct analogue of global aggregation weights in this
reading space: its failure is evidence that document-level conditioning,
not merely learned weighting, is the necessary ingredient. Judge
instability itself is well
documented~\cite{han2026swapping,gringras2026scaffolding}; we build on
the diagnosis rather than repeat it.
\paragraph{LLM-judge reliability and bias.} The judge-ecosystem
instantiation of Section~\ref{sec:judgeeco} reads against a canon:
MT-Bench and Chatbot-Arena established LLM-judge agreement with human
preference and catalogued the failure modes
\cite{zheng2023judging}; the Arena platform's methodology is the
provenance of our battle data \cite{chiang2024chatbot}; position bias
and its mitigation were analysed before judge ensembles became
common \cite{wang2023large}; panel-of-models evaluation is the direct
prior art for multi-judge composition \cite{verga2024replacing}; and
self-preference bias grounds the judge-error findings of our
validity decomposition \cite{panickssery2024llm}. The contested
share we measure is a complement to these agreement studies: it
prices pairwise panel instability, not accuracy.

\paragraph{Measurement-theory reading.} The contract axes are
facets in the sense of generalizability theory
\cite{cronbach1972dependability}: the measured dimension hierarchy
is a coarse D-study quantity, and the scoping map states the facet
conditions under which a census certificate generalizes; scoped
certificates are a quantified special case of argument-based
validity \cite{kane2013validating}. We keep the mapping implicit in
the body and flag it here for the measurement reader; the
judge-ecosystem section reports chance-corrected agreement
(Krippendorff $\alpha$) alongside its $\bar\tau$ figures
\cite{krippendorff2004reliability}.

\paragraph{Same-axis work.}
CONSISTRE~\cite{sun2026consistre} shares our data axis---rescoring
identical DocRE predictions under Re-DocRED---but asks a different
question: when does rule-based consistency repair pay off? It introduces
the $\Delta_{\mathrm{closure}}$ diagnostic (computable from annotations
alone) to predict repair benefit. Our work is complementary: we ask when
\emph{aggregation} (not repair) is stable, and our applicability
criterion (Section~\ref{sec:discussion}) plays the role that
$\Delta_{\mathrm{closure}}$ plays for repair. The two are composable:
a practitioner could run the closure diagnostic to decide whether to
repair, then our exchangeability check to decide how to aggregate the
resulting readings. Re-DocRED itself~\cite{tan-etal-2022-revisiting} (the gold-revision source for $g_R$)
is the ground-truth layer beneath both; our contract axis reads it as one
endpoint, not as a correction to be applied.

\paragraph{Annotator-side disagreement.}
A mature line models annotator disagreement---STABLEVAL defines stability as
expected Kendall's $\tau_b$ under annotator subsampling~\cite{bonagiri2026stableval};
NUTMEG separates systematic from noisy disagreement over annotator
subpopulations~\cite{ivey2025nutmeg}; a recent survey systematizes the
field~\cite{xu2026beyond}. These methods aggregate \emph{annotator labels} to
recover or stabilize ground truth; we aggregate \emph{evaluation readings}
induced by a controlled, exhaustive contract axis, where the variation to be
pooled is a systematic property of the evaluation protocol rather than
independent annotator noise, and the pooled object is a thresholded structural
decision, not a flat label. The survey's noted open challenge---task-induced
disagreement from task formulation and label schemas---is precisely the axis
we operationalize.

\paragraph{Conformal evaluation.}
Conformal prediction and its risk-control extensions
\cite{angelopoulos2024crc} provide distribution-free guarantees and
have been applied to evaluation-adjacent problems such as MT quality
estimation intervals~\cite{giovannotti2023cpmt} and prediction sets
for human decision aid~\cite{cresswell2024cpsets}. These works
calibrate uncertainty \emph{of predictions}; our Part~2 applies risk
control to a different object---the decision layer of an evaluation
reader trained on contract readings---and the guarantee itself is
cross-referential: the dial is calibrated on one evaluation reference
and certified on another, making reference exchangeability (our
protocol's step (a)) the validity condition. To our knowledge, no
prior work risk-controls the wrong-commit rate of a learned
evaluation reader.

\paragraph{Metric combination and robust rank aggregation.}
Learning to combine metric readings predates our work: regressive ensembles
predict quality from heterogeneous metric vectors~\cite{stefanik2021regressive},
and DiscoTK tunes global interpolation weights, perturbing ties away as
post-processing~\cite{joty2020discotk}. CoarsenRank attains distributional
robustness against annotator misspecification by raising a generative ranking
likelihood to a global exponent~\cite{pan2020coarsenrank}. All operate within
a single fixed contract, treat ties as noise or forbid them, and learn global
static weights; none targets invariance under contract switch. Tie-aware
differentiable ranking objectives~\cite{he2018hashing} supply the mathematical
machinery our soft-tie loss adopts---averaging over orderings of tied items
via soft histograms---repurposed from Hamming retrieval to contract-readout
stability. Item-response-theory audits of
benchmarks~\cite{land2026auditing} diagnose items against a fixed reference;
we fix the reference and vary the reading contract.

\paragraph{Certification with scope.}
Selective evaluation with guarantees certifies instance-level judge
reliability~\cite{jung2024trust}; conformal risk
control~\cite{angelopoulos2024crc} and its weighted extension under
covariate shift~\cite{tibshirani2019weighted} supply the machinery our
Part~3 tests, and beyond-exchangeability analysis~\cite{barber2023beyond}
predicts where unweighted certificates break. To our knowledge no
prior work scopes certificates by evaluation-contract concordance; the
five-step protocol of Part~3 fills exactly that gap, and its
certificate machinery (fixed-sequence
testing~\cite{bauer1991multiple} over exact binomial bounds) is the
same statistical core as the instance-level line applied to a
different object---pairwise orderings under contract swaps. Benchmark
agreement testing~\cite{perlitz2024bat} asks whether benchmarks agree;
we ask which orderings survive their disagreement.

\section{Problem formulation}
\label{sec:formulation}

\paragraph{Aggregation skepticism.}
In a recent position paper, Gilda and Gilda~\cite{gilda2026perishable}
document \emph{trust inflation}: averaging multiple evaluation signals
can make reported confidence exceed the reliability of the weakest
signal, and on HELM the top-five models under mean aggregation are
entirely disjoint from those under a weakest-link aggregation. Their
response is to stop aggregating---attach metadata (formality tier,
scope, expiration) to each score instead. We share their premise that
naive aggregation manufactures confidence, but draw the opposite
conclusion: the fix is not to abandon aggregation but to make it
auditable---per-document weights, marginalization over contracts, and
a decision layer whose wrong-commit rate would carry a
distribution-free guarantee when audit references are
exchangeable---a condition our preregistered application shows fails
exactly on contested pairs (Section~\ref{sec:part2}). Their HELM
evidence doubles as
third-party support for our degenerate-corner argument: aggregation
strategy changes conclusions, so the aggregation itself must be
preregistered and audited.

\paragraph{Contracts and readings.}
Let $M$ be a system (generator) and $\mathcal{C}$ a set of evaluation
contracts, each contract $c=(g,s,\tau)$ fixing a gold-standard version $g$, a
scoring function $s$, and a thresholding rule $\tau$. Following the
companion~\cite{lu2026evaluation}, the \emph{reading} of $M$ under $c$ is the
intra-minus-inter sentential gap of the contracted metric over the evaluation
documents $\rho(M,c)$; for the scoring axis we use the contracted column
(hit-any $A$ or label recall $R$) directly. The tuple is deliberately
open beyond $(g,s,\tau)$: Section~\ref{sec:discussion} shows the
reading function $\rho$ is itself an evaluative coordinate, and the
registry admits additional endpoint axes (the NA-margin of
Section~\ref{sec:part2}). What is closed is the requirement that
\emph{every} axis be declared and frozen before readings are
compared---openness is fine, undeclared freedom is not. The slicing
convention is
stated exactly: candidates are partitioned by sentence position---a pair
whose head and tail entities first occur in the same sentence is an
\emph{intra} candidate (group~0), otherwise \emph{inter} (group~1)---and
both strata's micro-$F_1$ are computed over \emph{all} $498$ documents
with no document-level subsetting. This differs from the document-subset
convention used by common follow-up evaluation codebases (e.g., the
thunlp/DocRED repository's later re-evaluation scripts and published
intra/inter tables), which compute the inter-side $F_1$ only over the
subset of documents containing at least one inter-sentential gold
relation---a convention the official evaluation script predates; our convention keeps every candidate in exactly one stratum,
so the gap decomposes additively across strata and no document is
silently dropped from either side. Predictions are frozen: all
contracts read the same prediction pool, so reading variation is attributable
to the contract alone.

\paragraph{Flip registry and tie declaration.}
A flip registry freezes, for every generator pair $(M_i,M_j)$ and
contract dimension (gold, scoring, threshold), the paired difference
$D_1=\rho(M_i,c_1)-\rho(M_j,c_1)$ under both contract endpoints with
bootstrap confidence intervals, computed by document-level paired resampling
($R{=}10{,}000$, percentile 95\%). A \emph{flip} is a sign change of the point
difference across the endpoints; 13 flips among 57 available checks.
A reader \emph{declares a tie} on a pair when the 95\% percentile interval of
its own resampled paired difference contains zero (touching counts as tied)---the
registry's frozen rule, which we adopt unchanged. The registry used
throughout is the companion submission's first assembly; our loader
reproduces it bitwise (Section~\ref{sec:protocol}), and
Appendix~\ref{app:registry} enumerates all 13 flips,
Table~\ref{tab:generators} lists the families, and
Appendix~\ref{app:contracts} enumerates the training contracts, so
this section is self-contained.

\paragraph{Dual objective.}
For a reader $\mathcal{R}$ producing scores $S_i=\mathcal{R}(M_i)$, we
evaluate (i) \emph{stability}: agreement of $\operatorname{sign}(S_i-S_j)$
with the held-out gold-side native ordering, registry coverage (ties declared
on flip pairs), false-tie rate (ties declared on stable pairs), and
hard-ranking fidelity (sign correctness on flips the reader declines to tie);
and (ii) \emph{quality}: Kendall's $\tau_b$ against reference orderings. The
goal is to move the coverage--false-tie frontier outward---higher coverage at
matched false-tie, or matched coverage at strictly lower false-tie---without
degrading quality anchors.

\paragraph{Design desiderata.}
Three commitments follow directly from this formulation. First,
\emph{exchangeable readings}: the contract set must read one quantity
under legal variation. The DocRE family is the candidate instance, but
its two golds are related by revision rather than sampled
independently---Re-DocRED exists precisely because the original
annotation is systematically incomplete---so exchangeability holds at
best \emph{conditionally}, away from the contested pairs where the
revision disagrees with the original. The protocol therefore tests the
premise rather than assumes it, on DocRE itself as well as on TACRED;
Part~2's registered failure (Section~\ref{sec:confirmv2}) is the
empirical content of that test. Admission to the \emph{registry} is
distinct from admission to the \emph{consensus}: the registry is a
multi-ledger audit structure whose axes need only be declared and
frozen, while the consensus pool---the training contracts---admits
only axes that pass the exchangeability screen. The NA-margin
endpoint illustrates the boundary: it enters the registry as an audit
axis but not the training contracts, and the per-stratum accounting
keeps its ledger separate. Second,
\emph{hedge or commit, never guess}: on contested pairs the reader should
either declare a tie or side with the contract consensus; committing
against every contract is the signature failure of single-contract
supervision, and our consensus objective is designed to make it
non-optimal. Third, \emph{price the ties}: because the all-tie corner
satisfies any coverage-only criterion, tie declarations must be evaluated
jointly with their false-tie cost---the Pareto reading that structures
every judgment in Section~\ref{sec:protocol}. Fourth,
\emph{preregister the readout}: since the consensus target and the
reference contract can disagree by construction
(Proposition~\ref{prop:tradeoff}), which criterion will count as
success---and which negative result will be reported as such---must be
fixed before the confirmatory run, not chosen after it; this commitment
is what separates a registered audit from a post-hoc narrative, and it
is the step most often skipped in informal evaluation practice.

\paragraph{Degenerate corner.}
A reader that declares every pair tied attains perfect coverage and zero
information. Coverage-only metrics reward this corner; our primary judgments
are therefore Pareto conditions (coverage \emph{and} false-tie), and we report
the all-tie corner explicitly as the null geometry of the frontier.

\section{Method}
\label{sec:method}

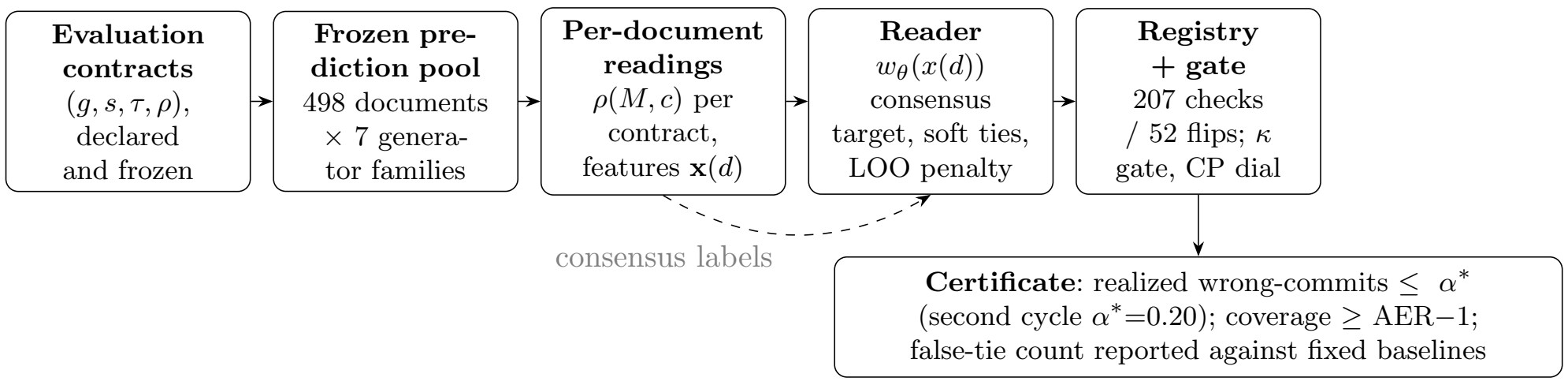
\begin{figure}[t]
\centering
\resizebox{\textwidth}{!}{%
\begin{tikzpicture}[every node/.style={font=\small},
  box/.style={draw, rounded corners, align=center, inner sep=4pt,
  minimum height=13mm, text width=2.5cm}]
\node[box] (c) {\textbf{Evaluation contracts}\\ $(g,s,\tau,\rho)$,
declared and frozen};
\node[box, right=2.5mm of c] (p) {\textbf{Frozen prediction pool}\\
498 documents $\times$ 7 generator families};
\node[box, right=2.5mm of p] (r) {\textbf{Per-document readings}\\
$\rho(M,c)$ per contract, features $\mathbf{x}(d)$};
\node[box, right=2.5mm of r] (a) {\textbf{Reader} $w_\theta(x(d))$\\
consensus target, soft ties, LOO penalty};
\node[box, right=2.5mm of a] (g) {\textbf{Registry + gate}\\
207 checks / 52 flips; $\kappa$ gate, CP dial};
\node[box, below=7mm of g, text width=8.0cm] (cert)
{\textbf{Certificate}: realized wrong-commits $\le\alpha^{*}$
(second cycle $\alpha^{*}{=}0.20$); coverage $\ge$ AER$-1$;
false-tie count reported against fixed baselines};
\draw[-Stealth] (c) -- (p);
\draw[-Stealth] (p) -- (r);
\draw[-Stealth] (r) -- (a);
\draw[-Stealth] (a) -- (g);
\draw[-Stealth] (g) -- (cert);
\draw[-Stealth, dashed] (r.south) to[bend right=30]
  node[pos=0.45, below left, font=\normalsize, text=gray]{consensus labels} (a.south);
\end{tikzpicture}}
\caption{The audit pipeline. Contracts and the prediction pool are
frozen first; readings and document features feed a document-conditioned
reader trained toward the contract consensus; the aggregated ranking is
audited against the registry, and the Part-2 gate layer turns reader
confidence into a risk-controlled commit decision whose realized error
rate is audited against a reference that never entered the gate}
\label{fig:pipeline}
\end{figure}
\subsection{Per-document readings and features}

Table~\ref{tab:components} summarizes the method's components, their
lineage, and each one's certification status before the details.

\begin{table}[t]
\centering
\caption{Method components, their lineage, and certification status.
The learning machinery is deliberately minimal; the contribution rests
on the protocol around it, and each component's evidential status is
stated where it is evaluated}
\label{tab:components}
\small
\begin{tabular}{p{2.8cm}p{2.9cm}p{2.2cm}p{3.2cm}}
\toprule
Component & Role & Lineage & Certification status \\
\midrule
Per-document contract weights (16-hidden MLP) &
document-conditioned weighting over contract readings &
new &
corpus-level collapse confirms conditioning is necessary \\
Contract-consensus objective (median-sign target) &
supervision signal &
new &
Prop.~1; fidelity $43/43$ (training sanity) \\
Soft-tie kernel with annealing &
differentiable tie handling &
adapted from tie-aware ranking objectives~\cite{he2018hashing} &
loss ablation: inert at registry granularity \\
Leave-one-contract-out penalty ($\lambda$) &
stability regularizer &
new &
ablation: the only active lever \\
Risk-controlled gate + CP dial (Part~2) &
decision layer over reader commits &
Risk-Controlling Prediction Sets (RCPS)~\cite{bates2021rcps} applied to reader decisions &
gate-layer J5$'$--J7$'$ (registered, second cycle) \\
\bottomrule
\end{tabular}
\end{table}

Fig.~\ref{fig:pipeline} overviews the pipeline: contracts and the
frozen pool feed per-document readings and features; the reader learns
document-conditioned weights toward the consensus; the gate layer turns
its confidence into certified commits. For each generator $M_g$ and
contract $c$ we compute, per document $d$, the
contracted column gap $r_c(d,g)=\text{col}_c(\text{intra})-\text{col}_c(\text{inter})$,
standardized within document across the pool to $z$-readings $\hat z_c(d,g)$.
The feature vector $x_d$ (7 classes, 18 columns) comprises: cross-generator
dispersion of document F1 and prediction rate; gold relation counts, type
counts, and cross-sentence gold fraction (via candidate--gold join); document
length; the contract reading vector; pre-aggregability statistics
(gap correlations across gold sides, near-zero-gap fractions---a reading-side
analogue of closure diagnostics~\cite{sun2026consistre}); near-threshold
density (share of candidates with maximum sigmoid probability in
$[0.4,0.6]$); and margin/arm proxies instantiating the companion's no-flip
margin guarantee. Features touching the held-out gold side are excluded from
the model input (leakage control).

\subsection{Document-conditioned contract weighting}
A shared trunk maps $x_d$ to logits over the $K$ training contracts,
separately for each generator:
\begin{equation}
w[d,g,k] \;=\; \operatorname{softmax}_k\!\bigl(W_2\,\phi(W_1 x_d + b_1) + b_2\bigr)_{g},
\qquad \sum_k w[d,g,k]=1 ,
\end{equation}
and the aggregated reading is the candidate-mass-weighted document mean of the
contract mixture:
\begin{equation}
S_g \;=\; \sum_d q_d \sum_k w[d,g,k]\,\hat z_{c_k}(d,g),\qquad
q_d = n^{\text{cand}}_d \Big/ \textstyle\sum_{d'} n^{\text{cand}}_{d'} ,
\end{equation}
with generators lacking a contract axis (the native-rule bridge) renormalized
over their available contracts. Uniform weights recover mean-of-$z$
marginalization; a weight collapse to one contract recovers the best-single
corner; both are reported as degenerate points of the same parameterization.

\subsection{Contract-consensus objective with soft ties}
For each generator pair let $\delta_c(i,j)$ be the pooled contracted-column
difference under contract $c$. The supervision signal is the
\emph{consensus}: $\sigma_{ij}=\operatorname{sign}(\operatorname{median}_c
\delta_c(i,j))$ and the strength $\kappa_{ij}$, the share of contracts
agreeing with the median sign (a raw agreement share, deliberately
not chance-corrected---it is not Cohen's $\kappa$). The quality loss is a strength-scaled
soft-margin hinge in softplus form,
\begin{equation}
\mathcal{L}_{\text{qual}} \;=\; \frac{1}{|\mathcal{P}|}\sum_{(i,j)}
\operatorname{softplus}\!\bigl(m\,\kappa_{ij}\,(1-t_{ij}) \;-\; \sigma_{ij}\,\Delta S_{ij}\bigr),
\end{equation}
where $\Delta S_{ij}=S_i-S_j$ and the \emph{soft-tie degree}
$t_{ij}=\max\{0,\,1-|\operatorname{median}_c\delta_c(i,j)|/\Delta\}$ follows the
tie-aware philosophy of averaging over tie orderings~\cite{he2018hashing}:
pairs whose contracts nearly cancel are neither forced apart nor collapsed---
their required margin melts as the kernel width $\Delta$ anneals during
training. This design matters: an earlier formulation supervised by the
single-contract $g_D$ sign provably caps the aggregator at reproducing $g_D$
(Section~\ref{sec:ablations}), whereas consensus supervision lets the reader
hedge on contested pairs and commit where contracts agree.

\paragraph{Consensus fidelity vs.\ contract fidelity: a trade-off.}
The objective optimizes fidelity to $\sigma_{ij}$, not to any single
contract. The two notions diverge exactly where contracts disagree, and the
price of the divergence is quantifiable.

\begin{proposition}[Consensus--contract fidelity trade-off]
\label{prop:tradeoff}
Fix a pair set $\mathcal{P}$ and a reference contract $c^\ast$ (e.g., the
held-out $g_R$ native reading), with per-pair signs $c^\ast_p$.
(i) A reader that is consensus-perfect on $\mathcal{P}$ (its hard-ranked
sign on every $p$ equals $\sigma_p$) has fidelity against $c^\ast$ exactly
equal to the consensus--$c^\ast$ agreement rate,
$\operatorname{fid}_{\mathcal{R}}(c^\ast)
=|\{p:\sigma_p=c^\ast_p\}|/|\mathcal{P}|$.
(ii) Perfect consensus fidelity and perfect $c^\ast$-fidelity are jointly
attainable if and only if $\sigma$ and $c^\ast$ agree on all of
$\mathcal{P}$.
\end{proposition}
\begin{proof}
(i) On each $p$ the reader outputs $\sigma_p$; a pair contributes an error
against $c^\ast$ iff $\sigma_p\neq c^\ast_p$, so the error count is
exactly $|\{p:\sigma_p\neq c^\ast_p\}|$.
(ii) If $\sigma$ and $c^\ast$ agree everywhere, the consensus-perfect
reader is $c^\ast$-perfect; if they disagree on some $p$, that pair is an
error against $c^\ast$ for every consensus-perfect reader.
\end{proof}

The bound is tight in the confirmatory run: consensus and the held-out
$g_R$ contract disagree on exactly two of the learned reader's eight
hard-ranked pairs, and the reader's $g_R$ fidelity ($6/8$) attains the
agreement rate exactly, while the $g_D$-native reading sits at the opposite
corner ($4/6$ consensus, $6/6$ against $g_R$). Proposition~\ref{prop:tradeoff}
thus converts the DocRE ranking-anchor failure
(Section~\ref{sec:experiments}) from a defect into a structural property:
the price of certified consensus fidelity equals the contract-disagreement
rate, and it is zero precisely when contracts are exchangeable in the
strong sense of always agreeing on contested pairs.

\subsection{Leave-one-contract-out stability penalty}
Stability is trained, not only measured: for each training contract $c_k$ the
leave-one-out reading $S^{(-k)}$ is recomputed with renormalized weights and
the pairwise sign concordance penalized,
\begin{equation}
\mathcal{L}_{\text{flip}} \;=\; \frac{1}{K}\sum_{k}
\frac{1}{|\mathcal{P}|}\sum_{(i,j)}
\operatorname{softplus}\!\bigl(-\Delta S_{ij}\,\Delta S^{(-k)}_{ij}\,/\,\epsilon\bigr),
\end{equation}
and the total loss is $\mathcal{L}=\mathcal{L}_{\text{qual}}+\lambda\,
\mathcal{L}_{\text{flip}}$. Training is full-batch Adam on CPU
($\sim$3{,}000 epochs, seconds per run), with all weights constrained
non-negative and contract-normalized by construction---without normalization
constraints, scale and reflection degeneracies make stability claims vacuous,
a lesson taken from the rating-stability literature~\cite{liu2025amelo}.

\subsection{Implementation and complexity}
The trunk is a two-layer network (16 hidden units, ReLU) producing $G{\times}K$
contract logits; weights are non-negative and contract-normalized by the
softmax, so no separate constraint step is needed. Training is full-batch Adam
(3{,}000 epochs, learning rate $10^{-2}$, weight decay $10^{-4}$), with the
tie-kernel width annealed linearly from $1.0$ to $0.3$ over the first 60\% of
epochs and the flip weight fixed at $\lambda{=}2$. Every reported run trains in
seconds on a single CPU core; the entire experimental suite of this
paper---including the $B{=}2{,}000$ bootstrap evaluations---completes in under
two CPU-hours, since all learning operates on per-document count vectors
rather than model forward passes. The feature table (498 documents $\times$ 18
columns) is computed once from the frozen pools; near-threshold density reads
candidate logits in a single streaming pass, with the rule-bridge generator
(excluded from threshold contracts by construction) contributing no continuous
scores. Feature standardization is fit on the full document set before
splitting; no held-out gold-side feature enters the model input.

\subsection{Baselines as degenerate points}
The harness evaluates, under identical protocols: \emph{uniform} (equal
contract weights; the AER analogue in reading space), \emph{consensus-median}
(median contract reading), \emph{AER (corpus form)} (the companion's
equal-weight marginalization of corpus-level $z$-scores over the same training
contracts), \emph{native} single-contract readings ($g_D$; $g_R$ as the
held-out reference), a \emph{corpus-learned} variant (one weight vector per
generator, no document conditioning), and a \emph{trivial-feature} variant
(document length and candidate count only).

\section{Evaluation protocol}
\label{sec:protocol}

\emph{Roadmap.} The protocol below is the audit-cycle form of steps 1--2 (Declare: contract axes frozen; Measure: the frozen readout table) together with the power arithmetic that Part~3 generalizes into the Scope step.

The confirmatory evaluation below instantiates steps (b)--(d) of the
audit protocol; the exchangeability check (a) is examined ex ante in
Section~\ref{sec:formulation} and its violation demonstrated on TACRED,
and the probes (e) are reported in Section~\ref{sec:discussion}.

\paragraph{Loading gate (bitwise).}
All learning consumes frozen per-document counts re-implemented by an
independent loader. As a hard gate, the loader must reproduce the companion's
registry bitwise: gaps, leverage, flip matrix, Kendall maps, top-5 robust
flips (57 checks, 13 flips), and the registry's tie-coverage statistics
(ens 3/13, union 5/13, floor 4/13 with the full false-tie suite); and the
canonical WMT19 marginalization numbers ($\tau_b$ .700/.7524, flip coverages)
must match to the last digit. Both gates pass; every table below therefore
shares the companion's arithmetic foundation.

\paragraph{Metrics.}
(i) \emph{Sign agreement} and $\tau_b$ against the held-out gold-side native
ordering (document-level split 350/148, five seeds; paired bootstrap over
documents, $B{=}1{,}000$--$2{,}000$). (ii) \emph{Registry coverage / false tie}
under the frozen tie rule. (iii) \emph{Hard-ranking fidelity}: on flip pairs
the reader declines to tie, does its sign agree with the held-out gold
standard and with the contract consensus? (iv) \emph{Frontier}: coverage and
false tie swept over confidence levels 50--99\%.

\paragraph{Transfer.}
TACRED~\cite{zhang-etal-2017-tacred}: 45 released systems, three gold variants ($G_0$ original, $G_1$
TACREV revision~\cite{alt-etal-2020-re-tacred}, $G_2$
Re-TACRED~\cite{stoica-etal-2021-retacred}), corpus-level readings only; training
contracts from the $G_0$/$G_2$ axes, $G_1$ held out; the frozen symmetric
registry (88 CI-status flips of 990 pairs on micro-F1, validated exact
against released gauges) supplies flip pairs; the metric is \emph{flip
fidelity}---the share of flip pairs on which the aggregated reading sides
with the revised gold $G_1$.

\paragraph{Reproducibility.}
All learning consumes frozen per-document counts whose loader is gated by
bitwise reproduction of the companion registry and of the published
marginalization anchors; every table in this paper carries its provenance
(confirmatory seeds 20261201--05 or an explicit exploratory-phase label)
and every figure names its source file. The released harness executes the
entire confirmatory suite---three testbeds, five seeds, judgment
criteria---as a single CPU command, writes a manifest with the judgment
outcomes, and refuses to run while the preregistration is not frozen. The entry point (\texttt{confirmatory\_run.py}) and its full
invocation are documented in the released artifact README; the run
writes its manifest to
\texttt{out/confirmatory/}.
A preregistration deviation log accompanies the release; it records, among
others, a corrected significance-arithmetic bug in the automated judge
(the criteria themselves were never altered). Its consequential entry is
\emph{amendment~3} (2026-11-07): a pooled reference-reading arithmetic
correction---the reference gap was computed as an intra-side difference
rather than the intra--inter gap---which re-ran the entire confirmatory
pipeline. The amendment's footprint is bounded and stated: the J1--J3
judgments are bit-identical before and after (verified against the
archived manifests), J4 flips and is reported as a registered revision,
and every table whose values moved carries an ``amendment~3'' caption
note; per-table values were re-reconciled against the archived
manifests, and a pre/post numeric diff of every moved statistic
accompanies the submission as supplement.

\paragraph{Preregistered confirmatory protocol.}
All exploratory numbers above derive from a designated exploratory phase with
disjoint seeds. Before the confirmatory run, a preregistration document froze
the judgment criteria: J1, hard-ranking fidelity strictly exceeding the native
reading (one-sided Fisher $p<0.05$); J2, false-tie count at or below every
fixed baseline; J3, coverage within one pair of fixed marginalization; J4,
quality anchors (DocRE Pareto tolerance, WMT parity $|\Delta\tau_b|\le0.02$).
The confirmatory run used five fresh seeds (20261201--05; WMT three), the
frozen training objective and run-selection rule, and the same bootstrap
($B{=}2{,}000$); deviations, if any, are recorded in a public amendment log
rather than by editing the frozen criteria. We report outcomes against this
table verbatim, including criteria that fail: a registered negative on an
underpowered registry is evidence about the registry, not a license to shop
for tests. Three scope qualifications are stated rather than implied.
First, this confirmation is \emph{seed-level}: the fresh seeds replicate
training stochasticity on a fixed prediction pool and registry; they do
not replicate data collection, and exploratory and confirmatory phases
share the same 498 documents and generator pool. Second, the three
preregistrations (v1, v2, v3) were frozen in-repository
\emph{without third-party timestamps}: their full text, the amendment
log, and the manifest hashes accompany the submission so that the
freeze points are independently checkable against the artifacts;
hash-pinned manifests and ledger timestamps are the freeze evidence,
and no external notarization is claimed. Third,
registry growth across cycles was adaptive by amendment; the
sensitivity analysis above quantifies what that dependence costs, and
future growth pre-freezes its candidate-axis list
(Section~\ref{sec:confirmv2}); and rank-level stability under hedging
(Bradley--Terry propagation over the arena's model pairs) requires
the arena's pair metadata, which is access-gated, so the battle-level
certificates here are the computable layer---the retroactive
decode-calibration probe on the frozen v1 registry and the rank-stability
analysis are registered for the next cycle. And so that the thesis itself is
falsifiable, three outcomes, stated here before any further data are
collected, would refute it: (i) if any fixed baseline achieved a
\emph{lower} false-tie count than the learned reader with McNemar
$p<0.05$ on a confirmatory seed set, the false-tie headline
collapses; (ii) if consensus aggregation on a TACRED-like
conflicting-revision pool produced positive, significant flip
fidelity, the applicability boundary collapses; (iii) if removing
contested pairs---the conditional-exchangeability position this
paper retreats to---also removed the DocRE gains, the conditional
retreat itself would collapse and the aggregation premise would be
void outright. Any of the three, observed, would be reported on the
front page. One scope note is owed on clause~(i): read literally, it
is satisfied by any \emph{non-hedging} reader, whose false-tie
count is zero by construction---the consensus-median challenger of
Section~\ref{sec:confirmv2} is exactly such a reader. The clause's
intended scope, carried by the abstract headline and the frozen
readout table, is the \emph{hedging} class at non-inferior
coverage; we report the literal reading here rather than amend the
frozen text. The literal reading's arithmetic: on the $44$ stable checks of the
confirmatory registry the consensus-median posts zero false ties
against the learned reader's seven---all seven discordances run one
way, and the exact McNemar on that $7$-vs-$0$ split gives
 $p{=}0.016$---so the clause as frozen has fired, and the abstract
headline survives only under the hedging-class scope. For symmetry, the J2 false-tie contrast
($7$ vs.\ $13$ of $44$) carries an unpaired binomial
$p{=}0.048$ (bootstrap-standard-error unpaired test; the paired
test is the v8 preregistration's first item).

\paragraph{Run protocol.}
Main protocol: the first released run per generator (fixed selection rule,
declared in advance). A cross-run-mean variant (the registry's own protocol)
is reported as sensitivity analysis; the two are not interchangeable
(Section~\ref{sec:robust}).

\section{Experiments}
\label{sec:experiments}

\emph{Roadmap.} This section instantiates steps 1--2 of Part~3's
protocol (Declare, Measure) on one ecosystem---the contracts are
frozen and the reversal record measured---and feeds step 5's hedging;
the decision layer of Part~2 instantiates step 4 (Certify) and its
failure modes.


This section mixes three provenance levels by design, and each
table or figure carries its level in the caption: \emph{confirmatory}
numbers (frozen criteria, seeds 20261201--05), \emph{post-hoc}
analyses (design validation, labeled as such), and \emph{exploratory}
numbers (exploratory-phase runs, labeled). The order is thematic, not
chronological; the confirmatory material proper is
Tables~\ref{tab:main}, \ref{tab:boot}, \ref{tab:registry}, \ref{tab:ablation}, and~\ref{tab:tacred}, and
Section~\ref{sec:confirmv2}.

\subsection{Main agreement results}
\label{sec:main}

\begin{table}[t]
\centering
\caption{[CONFIRMATORY: seeds 20261201--05; amendment-3 pipeline] Held-out gold-side agreement (full 498 documents; sign agreement
with the $g_R$ native \emph{pooled-gap} ordering over 21 generator pairs;
mean$\pm$sd over 5 confirmatory seeds where applicable; reference-corrected
pipeline, amendment 3). Dev values in parentheses are dev-split agreement. ``---'' marks readings that do not induce a corpus-level ordering over the 21 pairs, so $\tau_b$ is undefined for them.
The $g_D$-native and AER $\tau_b$ values coincide because both
bootstrap-mean orderings agree with $g_R$ wherever defined}
\label{tab:main}
\begin{tabular}{lcc}
\toprule
Reader & Sign agreement ($\uparrow$) & $\tau_b$ vs.\ $g_R$ ($\uparrow$) \\
\midrule
$g_D$ native (single contract)      & \textbf{0.810} & \textbf{0.905} \\
AER, corpus form (training contracts)& 0.714 & \textbf{0.905} \\
Consensus median (fixed)             & 0.571 & 0.143 \\
Uniform contract mixture (fixed)     & 0.381 & $-0.238$ \\
\midrule
Learned, corpus-level weights        & $0.524\pm0.000$ & --- \\
Learned, trivial features            & $0.590\pm0.023$ (dev 0.667$\pm$0.095) & --- \\
Learned, full features               & $0.581\pm0.047$ (dev 0.648$\pm$0.088) & $0.181\pm0.047$ \\
\bottomrule
\end{tabular}
\end{table}

\begin{table}[t]
\centering
\caption{[CONFIRMATORY: seed 20261201; amendment-3 pipeline] Paired bootstrap deltas of the learned reader (full features,
confirmatory seed 20261201, single-seed bootstrap; the five-seed
spread is in Table~\ref{tab:main}) against fixed baselines; CI95 in brackets.
Against the held-out $g_R$ ranking the learned aggregate tracks the
consensus-median baseline (by construction) and trails the single-contract
readings}
\label{tab:boot}
\begin{tabular}{lc}
\toprule
Comparison & $\Delta$ sign agreement [CI95] \\
\midrule
Learned $-$ $g_D$ native        & $\mathbf{-0.208}$ $\mathbf{[-0.286,\,-0.095]}$ \\
Learned $-$ AER (corpus)        & $-0.111$ $[-0.190,\,+0.000]$ \\
Learned $-$ uniform mixture     & $+0.115$ $[-0.190,\,+0.381]$ \\
Learned $-$ consensus median    & $+0.042$ $[-0.048,\,+0.191]$ \\
\bottomrule
\end{tabular}
\end{table}

Table~\ref{tab:main} and Table~\ref{tab:boot} give the corrected headline:
against the held-out $g_R$ ranking, the learned aggregate is
\emph{statistically indistinguishable from the consensus-median baseline it
optimizes} ($+0.042$, confidence interval (CI) contains zero),
trails the $g_D$ native reading
($-0.208$, CI excludes zero), and does not separate from the uniform
mixture. Its value claim is therefore \emph{not} gR-ranking maximization:
the reader buys the lowest false-tie rate (Table~\ref{tab:registry}) and
perfect consensus-side fidelity (Table~\ref{tab:registry}, last column) at
the price of tracking the consensus rather than any single contract's
ordering. Two fixed marginalization variants appear in this paper and must not be
conflated. The \emph{document-level} uniform mixture standardizes readings
within each document before averaging, which preserves within-document orderings but destroys
corpus-level comparability of the aggregated scores (the same failure
mode we later document and repair on
WMT19); it is reported as the degenerate-geometry reference. The
\emph{corpus-level} AER---the companion's published procedure and our
primary fixed comparator---standardizes once across generators. All
``improvement over uniform'' claims refer to the corpus-comparable
confirmatory contrast in Table~\ref{tab:boot}; against corpus-level AER
the learned reader is statistically indistinguishable.

That the
native reading is the bar---not a baseline to beat everywhere---is the
content of Section~\ref{sec:registry}: on the pairs that matter (the registry
flips), matching is not enough, and the learned reader separates there.

\paragraph{Confirmatory outcome.}
Under the frozen criteria (Section~\ref{sec:protocol}), the confirmatory
run returns: J1 \emph{fail} (fidelity $8/8$ vs.\ $4/6$, one-sided Fisher
$p{=}0.165$---the registry cannot certify this contrast at
$\alpha{=}0.05$); J2 \emph{pass} (false ties $7/44$, below every fixed baseline: AER $13/44$, $g_D$ native $19/44$, uniform $44/44$); J3 \emph{fail} by exactly one pair (coverage $5$
vs.\ non-inferiority floor $6$); J4 \emph{split}: the WMT parity anchor
passes exactly ($|\Delta\tau_b| = 0.000$), while the DocRE Pareto anchor
fails---the learned aggregate's generator-level ranking agrees with the
held-out $g_R$ native reading at $\tau_b{=}0.18{\pm}0.05$, far below the
frozen tolerance ($g_D$ native: $0.90$, threshold $0.85$). The aggregate
is by construction consensus-leaning (Proposition~\ref{prop:tradeoff}): it
follows the median contract
stance, which on generator-level orderings is closer to the
$g_D$-side view ($\tau_b{=}0.56$) than to $g_R$; the anchor failure is the priced trade-off, not a
defect. We therefore claim the
confirmed advantages---false-tie reduction and perfect consensus-side
fidelity on contested pairs---report the fidelity contrast as a
descriptive point estimate bounded by registry power, and record the
DocRE ranking-anchor failure as a property of consensus-leaning
aggregation, not of the readings.

J3 misses the non-inferiority floor by exactly one pair (the \texttt{atlop\_pe8}|\texttt{docunet} tie); the supplement (Online Resource~1) gives the full analysis under both loss functions and both references, which shows the deficit is decisiveness, not the consensus target.

A post-hoc unanimity gate ($\kappa_{ij}{=}1$, declaring ties only where all training contracts agree) closes the coverage gap at zero false-tie cost; its full analysis and the two-flip case study are in the supplement (Online Resource~1), and Section~\ref{sec:registryv3} returns it as the preregistered element it motivated.

\paragraph{Selection-effect controls.} Three post-hoc controls (forced ranking, selection-matched supersets, per-seed verdicts) are preserved in the supplementary material; the like-for-like result is the selection-matched analysis cited in the Confirmatory outcome above.
\subsection{Registry coverage, false ties, and the frontier}
\label{sec:registry}

\begin{figure}[t]
\centering
\includegraphics[width=0.92\linewidth]{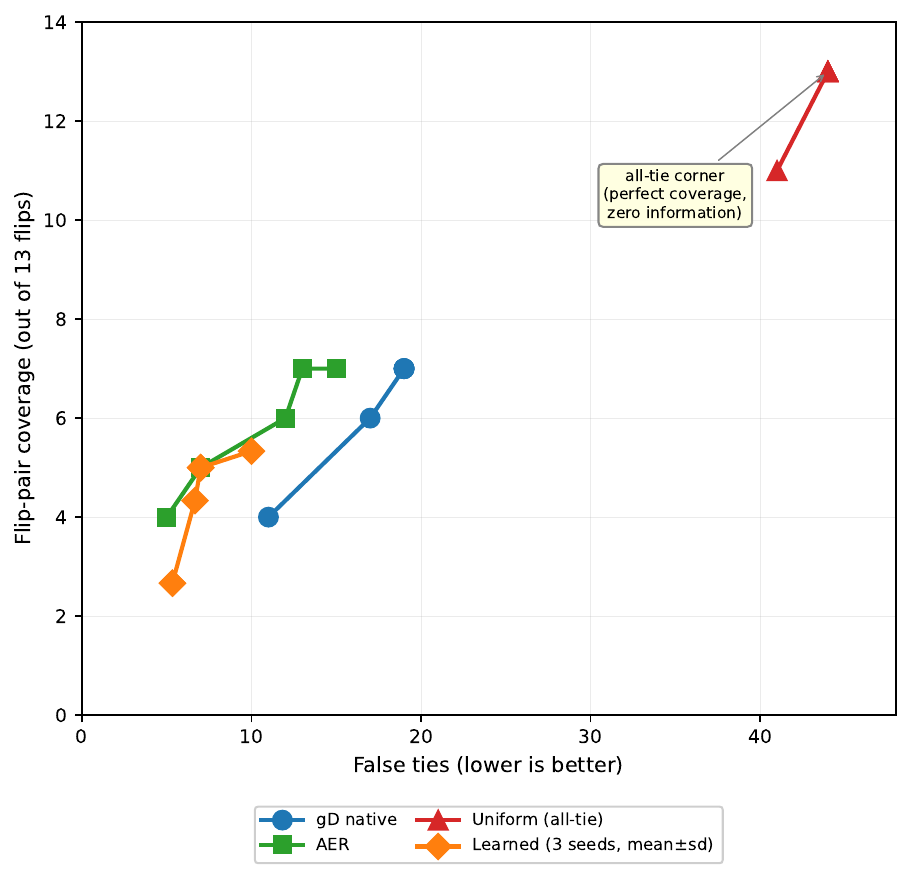}
\caption{[CONFIRMATORY: seeds 20261201--03 (three of five)] Coverage--false-tie frontier over tie-rule confidence
(confirmatory seeds 20261201--03, three of the five; the frontier
plateau is seed-stable across all five). The learned
reader holds a $5/7$ operating point from CI 70--95\% (seeds stable), matching
the informative end of fixed marginalization; the native reading is dominated
(7/19); the uniform reader collapses to the all-tie corner. The
confirmatory contrasts throughout use exact paired tests (McNemar,
Fisher) per the NLP significance-testing guidance
\cite{dror-etal-2018-hitchhiker,dietterich1998approximate}. Key
operating points: learned at CI 0.95 holds coverage 5 with 7 false
ties; AER at CI 0.95 holds coverage 7 with 13 false ties}
\label{fig:frontier}
\end{figure}

Fig.~\ref{fig:frontier} traces the coverage--false-tie frontier
this subsection quantifies.

\begin{table}[t]
\centering
\caption{[CONFIRMATORY: seeds 20261201--05] Frozen-registry reading outcomes at the 95\% tie rule
($B{=}2{,}000$). Coverage = ties declared on the 13 flip pairs; false tie =
ties on the 44 stable pairs; fidelity = sign correctness on the flip pairs the reader
hard-ranks, scored against the contract consensus (the training
target); against the held-out $g_R$ the same commits score $6/8$
(Table~\ref{tab:loss}); the consensus-median baseline is omitted from this table: as a fixed rule it hard-ranks every pair whose contract median is non-zero and ties otherwise---the decisiveness extreme, symmetric to the uniform reader's all-tie extreme; its registry-level challenger behavior is reported in Section~\ref{sec:confirmv2}}
\label{tab:registry}
\begin{tabular}{lcccc}
\toprule
Reader & Coverage & False tie & Hard-ranked & Fidelity \\
\midrule
$g_D$ native            & 7/13 & 19/44 & 6 & 4/6 \\
AER, corpus form        & 7/13 & 13/44 & 6 & 4/6 \\
Uniform mixture         & 13/13 & \textbf{44/44} & 0 & --- (degenerate) \\
\textbf{Learned (ours)} & 5/13 & \textbf{7/44} & 8 & \textbf{8/8} \\
\bottomrule
\end{tabular}
\end{table}

Three readings of Table~\ref{tab:registry}. First, the \emph{coverage--precision
positions differ}: the learned reader declares fewer ties on flips at the
lowest false-tie rate in the suite;
its confidence intervals are narrower, so it commits rather than hedges
generically. Second, the uniform reader's 13/13 with 44/44 false ties is the
degenerate corner of Section~\ref{sec:formulation}---coverage alone cannot
rank readers, and any primary judgment must be Pareto. Third, \emph{hard-ranking fidelity}---scored against the
consensus, the training target. When the learned reader declines a
tie on a flip pair, it ranks on the consensus-correct side on
\textbf{every} instance
($8/8$; the single contested pair in an earlier single-seed run,
\texttt{atlop\_pe8}$|$\texttt{kddocre}, is ranked toward the held-out gold
standard where the native reading is wrong). Native and corpus-AER readers,
by contrast, hard-rank flips at 4/6---their commitments on contested pairs
inherit one contract's bias. Against the held-out $g_R$, however, the
learned reader's $8$ commits score $6/8$ (Table~\ref{tab:loss}): it is
\emph{not} better than the fixed baselines on held-out fidelity, and
its confirmed advantages are the false-tie count and consensus fidelity. The learned reader's policy is, in effect,
\emph{hedge or side with the consensus}: it never hard-ranks against it.

\subsection{Ablations}
\label{sec:ablations}

\begin{table}[t]
\centering
\caption{[CONFIRMATORY: seeds 20261201--05 (post-hoc re-run)] Component ablations (sign agreement vs.\ the held-out $g_R$
ordering, full set / dev split; confirmatory seeds 20261201--05;
reference-corrected pipeline, amendment 3)}
\label{tab:ablation}
\begin{tabular}{lcc}
\toprule
Variant & Full 498 & Dev 148 \\
\midrule
Learned, full features            & $0.581\pm0.047$ & $0.648\pm0.088$ \\
Learned, trivial features (length, candidates) & $0.590\pm0.023$ & $0.667\pm0.095$ \\
Learned, corpus-level weights (no document conditioning) & $0.524\pm0.000$ & $0.514\pm0.097$ \\
\bottomrule
\end{tabular}
\end{table}

Under the corrected reference the ablation picture sharpens on one axis and
flattens on another (Table~\ref{tab:ablation}). Corpus-level weights still
collapse to a deterministic degenerate solution ($sd{=}0$ on the full set,
lowest on dev): document conditioning is necessary. The trivial-vs-full
feature contrast, however, no longer separates on $g_R$-side agreement
(the trivial and full variants sit within a point on both splits;
Table~\ref{tab:ablation}): with the reference
ranking fixed, the learned feature set is not load-bearing for tracking
$g_R$. Its load-bearing role, where one exists, must therefore be sought in
the consensus-side metrics---registry fidelity and false-tie rate
(Table~\ref{tab:registry})---not in gR-ranking skill. An earlier formulation
supervised by the $g_D$ sign alone plateaued exactly at the native reading's
performance (paired bootstrap CI containing zero, point estimate slightly
negative): single-contract supervision caps the aggregator at copying that
contract, which motivates the consensus objective. Leave-one-contract-out
analyses indicate the gold-axis contract is the heaviest information source,
consistent with the companion's leverage theory.

The loss-component ablation (Table~\ref{tab:loss}) shows the soft-tie kernel inert at registry granularity and the flip weight $\lambda$ the only active lever; the full per-variant narrative is in the supplement (Online Resource~1).

\begin{table}[t]
\centering
\caption{[POST-HOC: seed 20261201] Loss-component ablation (registry metrics; fidelity and sign
against the held-out $g_R$ ordering; post-hoc run on seed 20261201
(the seed value coincides with confirmatory seed~1 by convention; this
table is a post-hoc analysis and no confirmatory claim rests on it);
reference-corrected pipeline, amendment 3; $\lambda$ swept on a
log-spaced grid spanning the operative range). The soft-tie kernel and its
annealing do not move registry outcomes at this granularity; the active
lever is the flip weight $\lambda$}
\label{tab:loss}
\begin{tabular}{lcccc}
\toprule
Variant & Coverage & False tie & Fid. ($g_R$) & Sign ($g_R$) \\
\midrule
baseline($\lambda$=2) & 5/13 & 7/44 & 6/8 & 12/21 \\
$\lambda$=0 & 4/13 & 5/44 & 7/9 & 13/21 \\
$\lambda$=0.5 & 5/13 & 7/44 & 6/8 & 13/21 \\
$\lambda$=8 & 5/13 & 9/44 & 6/8 & 11/21 \\
no anneal(D=1.0) & 5/13 & 7/44 & 6/8 & 12/21 \\
no soft tie(t=0) & 5/13 & 7/44 & 6/8 & 12/21 \\
mean target & 5/13 & 7/44 & 6/8 & 12/21 \\
gD target(single-contract) & 5/13 & 7/44 & 6/8 & 12/21 \\
\bottomrule
\end{tabular}
\end{table}

\paragraph{Leave-one-generator-out.} LOGO controls (post-hoc) show held-out sign agreement $29/42$ ($69\%$) against in-sample $95\%$; the full fold table is in the supplementary material.
\subsection{Robustness}
\label{sec:robust}

\paragraph{Robustness.} Seed spread is zero on registry outcomes (deterministic convergence across all five confirmatory seeds); candidate-mass sensitivity and run-protocol detail are in the supplementary material.
\paragraph{Run protocol sensitivity (declared).}
Averaging per-document counts across each generator's released runs---the
registry's own protocol---shrinks document-level readings toward the mean and
degrades the aggregator markedly (coverage $2/13$; consensus-side fidelity
$9/11$ preserved, but $g_R$-side ranking collapses: match $3/11$,
$\tau_b$ $-0.33$); the confirmatory seed reproduces these values exactly,
indicating a structural property of the reading protocol rather than
training stochasticity. We therefore preregister the single-run selection rule as
the main protocol and report the cross-run mean as a sensitivity analysis;
per-run variance treatment is an identified improvement axis. Reading
protocol choices at this granularity are part of the contract, echoing the
companion's central claim.

\subsection{Transfer and the applicability boundary}
\label{sec:transfer}

\begin{table}[t]
\centering
\caption{[CONFIRMATORY: seeds 20261201--05] TACRED transfer (45 systems; 88 flip pairs, micro-F1, $G_1$ held
out). Flip fidelity = share of flip pairs sided with the revised gold $G_1$;
Wilson 95\% CI where applicable; learned rows are mean$\pm$sd over 5 confirmatory seeds (20261201--05). The $G_0$ row is ``definitional'' because $G_0$'s micro-fidelity is zero for every reader by construction: $G_0$ is the original gold, and no reader trained on revisions can recover its exact label set}
\label{tab:tacred}
\begin{tabular}{lccc}
\toprule
Reader & Micro fidelity & Macro fidelity & Overall (micro) \\
\midrule
$G_0$ native            & 0.000 (definitional) & --- & 0.871 \\
Consensus median        & 0.239 & 0.671 & 0.869 \\
Uniform (AER analogue)  & 0.261 $[0.181,0.362]$ & 0.693 & 0.871 \\
Learned, corpus weights & \textbf{0.275$\pm$0.005} & 0.680$\pm$0.005 \\
\textbf{Learned, conditioned} & 0.268$\pm$0.009 & \textbf{0.695$\pm$0.005} \\
\bottomrule
\end{tabular}
\end{table}

The method transfers operationally: on TACRED the learned readers sit at or
above the uniform baseline on every column, with macro-column differences
within CI overlap (Table~\ref{tab:tacred}), and the
frozen registry (88/990 flips) validates exactly against released gauges.
But the margins are small, and the reason is structural: the training-side
golds $G_0$ (original) and $G_2$ (Re-TACRED, pruned test set, disjoint label
revisions) \emph{disagree with each other}---the pooled ``contracts'' are
conflicting gold revisions, not exchangeable readings of one quantity, so
their consensus cannot reconstruct the held-out $G_1$ direction (micro
fidelity of \emph{all} readers clusters near $0.24$--$0.27$, far below the
macro column and far below DocRE levels). We elevate this negative result to
an \emph{applicability criterion}: consensus aggregation presumes the
contract set reads one quantity under legal variation; where gold revisions
supersede rather than re-read, the premise fails and the advantage vanishes.
DocRE's contract family satisfies the premise; TACRED's revision
structure does not. The legality criterion, restated from the
companion's analysis: a gold revision is a \emph{legal} variation when
it re-reads the same candidate pairs under the same task
definition---DocRED ${\to}$ Re-DocRED repairs and completes
annotations while leaving the candidate universe unchanged---and
\emph{illegal} when it re-adjudicates under a changed task:
Re-TACRED prunes thousands of entity pairs and relabels, so its
``readings'' are of a different quantity, not corrected readings of
the same one.

\paragraph{WMT19~\cite{barrault-etal-2019-wmt19}: parity on an exchangeable metric axis.}
WMT19 supplies the complementary case: three metric readings of one
quantity over 2{,}000 segments, an axis that satisfies the exchangeability
premise by construction. The learned reader achieves parity without harm
($\tau_b$ $0.683$ de-en and $0.771$ zh-en against the official human
anchor, identical to uniform marginalization and the best single metric,
reproducing the companion's published anchors and validating the loading
gate end-to-end). The full segment-level analysis is in the supplement
(Online Resource~1).


\section{Discussion and limitations}
\label{sec:discussion}
\label{sec:limitations}

\paragraph{What is learned.}
The aggregator does not beat the native reading everywhere---it is not
designed to. Its value is concentrated exactly where declarative evaluation
is weakest: contested pairs. There it converts an accepted bad trade-off
(either inherit one contract's bias, or tie everything) into a third policy:
hedge on contested pairs, and when committing, commit with the contracts
rather than against them. The $8/8$ vs.\ $4/6$ fidelity contrast and the
lowest-in-suite false-tie rate are two faces of this policy. The
component decomposition replicates under the confirmatory seeds:
document conditioning remains necessary (corpus-level weights again
collapse to a deterministic solution, $sd{=}0$), the trivial-feature
ablation shows the feature set is not load-bearing at registry
granularity (trivial $\ge$ full on both splits; Section~\ref{sec:ablations}),
and the uniform-mixture comparison no longer separates from zero
under the corrected reference ($+0.115$ $[-0.190,+0.381]$); the
pre-correction advantage ($+0.39$ $[+0.10,+0.71]$) was an artifact of
the reference arithmetic and is retracted (amendment~3). The same
verdict holds at the decision layer, and we state it as a finding:
twenty of the twenty-three direction-matched commits carry bootstrap
confidence $1.000$, leaving the calibration dial a single unit of
purchase---at $t^{*}{=}0.670$ it withdraws exactly one commit
(\texttt{docunet}|\texttt{kddocre}, confidence $0.647$) whose
direction is correct under the held-out revision, so the wrong-commit
count is unchanged. On this pool the learning and the calibration are not
load-bearing for the point estimate; what they buy is the certificate's
form. The natural next cycle preregisters the unanimity gate---the
learning-free rule that passes the same criteria---as a first-class
component, which is the honest form of this finding.

\paragraph{An ex-ante exchangeability check.}
The boundary condition must be screened ex ante: the mean inter-contract
Spearman $\rho$ is $0.83$ on DocRE, $0.89$ on TACRED, and $0.96$/$0.99$ on
WMT19---a necessary but not sufficient screen, because TACRED's healthy
global $\rho$ hides the conditional break on contested pairs. On the
consistent population (the training-contract contested set, $19/91$ DocRE
generator pairs) the $\kappa$ level is ${\approx}0.5$ by construction and
carries no cross-testbed signal; what transfers ex ante is the contested
density ($21\%$ DocRE vs.~${\approx}0$ WMT19). The $\kappa^{*}{=}0.75$
cross-testbed screen is accordingly withdrawn, the full derivation is in
the supplement (Online Resource~1), and the 2$\times2$ deployment
table below is reported as a post-hoc framing device.

\begin{table}[t]
\centering
\caption{The deployment 2$\times$2: exchangeability of the contract
set against the cost of flips. DocRE sits in the upper-left cell
conditionally (Section~\ref{sec:method}); the cell's deployment rule
is the learned reader with contested pairs hedged. Introduced after
both failures; reported as a post-hoc framing device
(Section~\ref{sec:confirmv2})}
\label{tab:decide}
\small
\begin{tabular}{p{2.6cm}p{5.4cm}p{3.4cm}}
\toprule
 & \multicolumn{1}{c}{Flip cost high} & \multicolumn{1}{c}{Flip cost low} \\
\midrule
Exchangeable & \textbf{Learn}, hedging contested pairs (DocRE,
conditionally) & Report parity (WMT) \\
Non-exchangeable & Single gold (--) & Report as-is (--) \\
\bottomrule
\end{tabular}
\end{table}

\paragraph{The reading function is itself a contract dimension.}
The paper defines the reading as the intra--inter gap
$\Gamma = F_1(\text{intra}) - F_1(\text{inter})$, following the
companion's sliced-gap formalization. But the choice of reading
function is itself undeclared evaluative freedom. Computing the same
contract switch ($g_D \to g_R$) under three reading functions yields
dramatically different flip counts and system rankings: the gap reading
produces 4 flips, while the inter-only and intra-only readings produce
8 each. The Kendall $\tau$ between the gap ranking and
the inter-only ranking is $0.333$---they measure fundamentally
different things. The mechanism is visible in the data: X2, a
multi-label model, ranks fourth by gap (its intra-F1 is high) but last
by inter-only F1 (0.13 vs.\ 0.37--0.53 for other systems). An
evaluator who reports the gap implicitly claims that the
intra--inter \emph{difference} is the quantity of interest; one who
reports inter-F1 alone makes a different claim. Neither is wrong, but
the choice must be declared---and the flip registry is conditional on
that declaration---so every registry must declare its reading
function $\rho$ alongside its other axes, and the combination
explosion (axes $\times\rho$) is bounded by the declaration
discipline, not by the taxonomy. This finding suggests that the full contract space
is larger than $(g, s, \tau)$: the reading function $\rho$ itself is
a coordinate that evaluators implicitly fix without reporting.

\paragraph{Registry size.} The released power simulation quantifies the registry-size ceiling ($n{\ge}20$ flips for 85\% power); see supplementary material for the full analysis.
\paragraph{Two flips in practice.} Two illustrative flip cases (threshold-axis and gold-axis) with their full reading trails are in the supplementary material.
\paragraph{Applicability.}
Consensus aggregation is not a universal solvent, and the three testbeds
now triangulate its applicability. The deployment question for any pooled
reading is twofold: are the members reading the same quantity
(exchangeability), and does ranking error on contested pairs carry real
cost (flip density)? DocRE answers yes to both and shows the largest
gains (perfect hard-ranking fidelity at the lowest false-tie rate); TACRED
answers no to the first---its training-side golds are conflicting
revisions---and the advantage vanishes with all readers trapped near
chance fidelity; WMT19 answers yes to the first and no to the second---
exchangeable metrics with no revised-gold direction---and yields exact
parity with fixed marginalization at the anchor. The learned aggregator
therefore neither always helps nor is inert: it helps exactly where
declarative aggregation was already known to be fragile.

\paragraph{Contract coverage.} The three uncovered candidate axes (Ign-F1, NA-handling, dev-leak) are enumerated in the supplementary material with their test timing.
\paragraph{Deployment cost.} Design-time screening costs CPU minutes; per-commit integration seconds; the full lifecycle table and per-stage measured costs are in the supplementary material.
\paragraph{Practical implications.}
For DocRE researchers reporting ablation tables: (1) if the margin
between two systems is smaller than the combined contract arms, declare
a tie---the comparison is not robust (the deployment reading of the
post-hoc 2$\times$2 table, Table~\ref{tab:decide}; a framing device,
not a validated rule); (2) if you must commit, check
that your ranking agrees with the direction supported by the majority
of available contract readings; (3) before deploying any evaluation
pipeline, verify exchangeability (do the contract readings correlate
across systems?) and estimate the registry power (can your flip count
certify the contrasts you care about?); and (4) run a \emph{decode-calibration robustness probe} before admitting any new contract
endpoint into a registry: re-determine flip flags under a calibrated
decode (per-relation prior-matched thresholds), and count only flips
that survive both decodes as robust. In our expansion, all nine
threshold-axis flips introduced by the new generators evaporate under
this probe---they are artifacts of the uncalibrated reference
decode---while every non-threshold flip survives
(Section~\ref{sec:discussion}). For benchmark maintainers: the
power simulation indicates that a flip registry below ${\sim}20$ pairs
cannot certify even perfect-vs-two-thirds contrasts; size
accordingly. We state the uncomfortable sequencing honestly: the
preregistration (v1) was frozen \emph{before} this power analysis
existed, and its J1 contrast therefore ran at ${\sim}44\%$ power---an
inconsistency we report rather than retroactively repair. Every
subsequent cycle incorporates the arithmetic (the v2 stop-gate of 35
flips and the v3 operating-point choice are both power-derived), which
is precisely why the cycles exist.
The companion's released power analysis provides concrete sizing: for
an effect of $0.02$, a DocRE evaluation needs $515$--$1244$ documents
(depending on the contrast) to reach $80\%$ power; for an effect of
$0.04$, $129$--$311$ documents suffice.

For registry designers assembling a preregistered evaluation registry
of their own, the five steps condense to a checklist: (1)\
\emph{exchangeability first}---verify that the candidate contracts read
one quantity before pooling them (Section~\ref{sec:discussion}, the
$2\times2$ test); (2)\ \emph{size before freezing}---run the power
simulation at the planned flip count and expand the generator pool
until the target contrasts are certifiable
(Section~\ref{sec:discussion}); (3)\ \emph{freeze the readout
table}---criteria, tolerances, and the negative-result branch, before
any outcome exists (Section~\ref{sec:protocol}); (4)\ \emph{probe
new endpoints}---every added contract endpoint passes a decode-calibration robustness probe, and only dual-decode-stable flips enter
the robust registry (Section~\ref{sec:discussion}); and (5)\
\emph{preregister decision rules}---tie gates, hedging thresholds,
and their semantics are declared ex ante, never fitted to outcomes
(Section~\ref{sec:part2}).

\paragraph{Limitations.}
(i) \emph{Registry power.} The frozen registry contains 13 flip checks and
at most 21 generator pairs; exact tests on contingency tables of this size
have sharply bounded power. A perfect $8/8$ versus $4/6$ fidelity contrast,
for instance, yields a one-sided Fisher $p$ of $0.165$: the registry cannot
certify such contrasts at $\alpha{=}0.05$ no matter which reader produces
them. Confirmatory claims are therefore anchored where power exists (false
tie counts, parity deltas, coverage non-inferiority margins) and reported
descriptively where it does not; replication on a larger frozen registry is
the natural remedy, and the released harness makes such replication
mechanical. A simulation makes the requirement concrete: with the
observed effect profile (perfect vs.\ two-thirds fidelity) and an exact
Fisher test at $\alpha{=}0.05$, a registry of $n{=}10$ flip pairs per
reader achieves only $44\%$ power; $n{=}20$ reaches $85\%$; $n{=}30$
reaches $99\%$. The current registry's 13 flips (with at most 8 and 6
ranked by the two readers) sit squarely in the underpowered zone.
Doubling the frozen flip count to ${\sim}40$---achievable by the
companion's own protocol applied to two additional contract pairs or a
modestly larger generator pool---would clear the $80\%$ threshold.
The companion's own power analysis (released as
\texttt{power\_analysis.json}) shows that this is systemic: the
minimal detectable effect at 80\% power ranges from $0.017$ to $0.031$
across the companion's three preregistered comparisons, requiring
$361$--$1244$ documents for an effect of $0.02$---far above the 498-doc
evaluation set. The power deficit is a property of the evaluation
population size, not of the aggregation method. We
release the power-simulation script so that registry designers can size
their audits before freezing them.
(ii) The registry is one frozen snapshot; 21 generator pairs bound statistical
resolution (single-pair differences move agreement by $0.048$), which is why
our primary judgments are Pareto conditions and fidelity contrasts rather
than small aggregate deltas. (ii$'$) \emph{Phase provenance.} Tables carry per-table provenance: the
main-agreement, registry, TACRED, ablation, bootstrap, and frontier
figures all report the confirmatory pipeline (seeds 20261201--05)
under amendment~3, as stated in their captions; genuinely exploratory
or post-hoc numbers are labeled as such where they appear. Deviations
from the frozen judgment standard are recorded in the public amendment
log rather than silently applied. (iii) Run-protocol sensitivity is real and
declared; per-run variance-aware aggregation is future work. (iv) TACRED
offers corpus-level readings only, so the document-conditioned form reduces
to a reading-conditioned corpus variant there. (v) The exact tests treat the 39 registry units as independent, but
pairs sharing a generator family are correlated (five families are
single-seed); a cluster-robust sensitivity---block bootstrap over
families---is a registered direction for future cycles, and until
then the exact $p$-values should be read as nominal. (vi) The
aggregator is
evaluated on ranking stability, not on downstream task selection; connecting
stable readings to better model-selection decisions is a natural next step.

\section{Part 2: registry v3 and the conformal decision layer}
\label{sec:part2}

\emph{Roadmap.} Part~2 instantiates step 4 (Certify) for one
ecosystem---including the registered negatives that become Part~3's
failure spectrum---and step 5's gate hedging.

Part~1 closed with three lessons, each of which becomes a
\emph{preregistered element} of the protocol's second application:
decode artifacts motivate a mandatory calibration probe per new family;
the coverage gap being a decision-rule property motivates a calibrated
decision layer in place of a fixed tie rule; and the power analysis
motivates growing the registry until the guarantee can bind. This
section is \emph{preregistered} (prereg\_v2.md): the generation
protocol and the decision-layer specification below were fixed before
any conformal outcome was computed, and its numeric results are added
only after the frozen confirmatory run.

\subsection{Registry v3}
\label{sec:registryv3}
The pool grows from nine to fourteen generator families: the seven
frozen families (ATLOP variants, DREEAM~\cite{ma-etal-2023-dreeam},
DocuNet~\cite{chen-etal-2021-docunet}, KD-DocRE~\cite{tan-etal-2022-document},
P3M, and the multi-label x2), DeBERTa-v3-base and Longformer-base (three seeds
each), and five newly trained families under the identical ATLOP
protocol~\cite{zhou-etal-2021-atlop} (RoBERTa-large, BERT-large-uncased, XLM-RoBERTa-large,
ELECTRA-base, mDeBERTa-v3; single seed each). Four instruction-tuned
LLM families (Qwen3-8B, GLM-4-9B, Llama-3.1-8B, Yi-1.5-9B; constrained
decoding over the same candidate universe, Wikidata-grounded relation
schema) enter \emph{conditionally} on a degeneracy gate ($\ge$500
predictions, $\ge$90\% document coverage, precision $>$2\%---a
conservative admission rule whose role is to keep corrupt predictions
out of the registry, not to grade the models). All four fail it, and
we report the failure as a \emph{harness-bound exclusion}, not a
finding about LLM relation extraction: published zero-/few-shot
LLM-DocRE results sit in the tens of F1 points, far above anything
this pipeline produces, and the failure signatures point inward---three
families massively over-generate (${\approx}34\times$ the gold pair
count at single-digit true positives), while GLM-4-9B collapses to
a near-empty prediction set;
opposite signatures of this kind indicate schema-grounding or
entity-pairing pathologies in the constrained-decoding harness itself
(per-family diagnostics in Appendix~\ref{app:llmdiag},
Table~\ref{tab:llmdiag}). The families
are therefore excluded as incomparable under this harness (recorded
amendment), and no claim about LLM capability is made or implied.
Three candidate backbones (DeBERTa-v3-large,
ELECTRA-large, Longformer-large-4096) collapsed under the sanity gate
and were excluded before their readings entered any registry
(registered amendments). Every admitted family passes the same
pipeline as Part~1: dual-decode evaluation, mirror integration with
the bitwise 57-check validation gate, and the calibration probe
(threshold-axis records survive only if they persist under the
prior-matched decode; all 42 new-family threshold records fail the
probe and are excluded). The registry additionally admits one new
contract endpoint by registered amendment, recorded after the initial
stop-gate shortfall and before any conformal outcome was computed: the
NA-margin axis (spurious-pair density on goldless candidate pairs,
intra minus inter). The endpoint is motivated from evaluation first
principles, not from the gate: how a protocol treats predictions on
goldless candidate pairs is a genuine protocol disagreement in
DocRE---the official scripts fold them into the false-positive mass,
while gold-pair-subset analyses ignore them entirely---so the density
at which a system hallucinates relations on semantically unrelated
pairs is a real property that conventional micro-F1 averages away,
and naming it an axis is the paper's own thesis (undeclared evaluation
choices are first-order decisions) applied to itself. Its readings
are near-orthogonal to the F1 endpoint ($\rho{=}-0.03$ across
families) and its 23 robust flips supply the heterogeneous flip mass
the stop-gate demands. A post-hoc decomposition prices that mass
honestly (supplement: \texttt{na\_margin\_flips.json}): across
all $69$ flipped checks---$47$ CI-excluding, of which the $23$ from
registry-v3 families are the registry's robust flips (the remainder
involve two audit-side families outside the registry)---the axis's
ordering is the exact reverse of the $F_1$ ordering (point
anti-symmetry $69/69$), so the flips carry definitional inversion
mass rather than pairwise instability; the stop-gate margin is
reported with and without the axis. The
assembled registry v3 (totals in Table~\ref{tab:pops}) passes the
stop-gate with margin ($52 \ge 35$): frozen 57/13; new-family
scoring and gold axes 98/8; $g_R'$ axis 29/8 ($g_{R'}$ being the
Re$^2$-DocRED second revision~\cite{heng-etal-2026-re2} of the same
498 development documents); NA-margin 23/23. The
conformal unit is the deduplicated flip pair: the $52$ flip checks
collapse to $39$ distinct pairs, of which $10$ involve a family
without a $g_R'$ reading and are excluded fold-wise, leaving $29$
$g_R'$-comparable units ($39$ remain judgable under $g_R$ labels).
The split is deterministic and fully specified: pairs sorted by
name-lexicographic order, even positions to calibration and odd to
test ($20/19$ overall; gated fold sizes $13/11$ and $13/10$), fixed
before any conformal outcome and never re-split. Confidence values
use the five-seed averaged bootstrap difference distribution with
$B{=}2{,}000$ resamples (seed 20261112); a definitions box for the
conformal vocabulary and the full per-unit record appear in
Appendix~\ref{app:perunit}.

Training hyperparameters follow the shared ATLOP protocol of Part~1
(30 epochs, linear warmup schedule, native AMP, seed 20261201): max
sequence length 512 and batch size 8 for RoBERTa-large,
BERT-large-uncased, XLM-RoBERTa-large, ELECTRA-base, and mDeBERTa-v3;
1024/8 for DeBERTa-v3-base; 2048/2 for Longformer-base under the
chunk-free compatibility model. Sanity gates per family: canonical
coverage 498/498 and true-positive mass in the established family
range (all seven admitted families passed; realized native-decode TP
counts 3,148--3,546); every family additionally passes the mirror
validation gate (57 frozen checks and 13 flips reproduced bitwise) and
the calibration probe before its contract records enter the robust
check set.

\subsection{The conformal decision layer}
\label{sec:conformal}
The fixed tie rule of Part~1 is replaced by a calibrated dial. For each
flip pair the reader's five-seed averaged bootstrap difference
distribution yields a confidence
$\mathrm{conf}(i,j)=\Pr\{\operatorname{sign}\Delta^{(b)}
=\operatorname{sign}\Delta\}\in[0.5,1]$, and the reader commits
iff $\mathrm{conf}\ge t$. The threshold is chosen by risk control
(RCPS~\cite{bates2021rcps}): the smallest $t$ whose calibration-fold wrong-commit rate has a
Clopper--Pearson upper bound $\le\alpha$ at confidence $1-\delta$
($\delta{=}0.1$, the conventional 90\% risk-control confidence,
frozen with the v2 specification). Because the reader is trained on the consensus, a
consensus-labelled calibration would be circular; the labels therefore
come from reference $g_R$ while the realized error rate is reported on
the held-out reference $g_{R'}$ (and symmetrically)---a
\emph{cross-reference} design whose validity condition is exactly the
reference exchangeability that step (a) of the protocol checks, and
which degrades to a conditional statement when references conflict.
On synthetic exchangeable pairs the construction controls the
population wrong-commit rate at $\alpha\in\{0.05,0.10,0.20\}$ in
$200/200$ Monte-Carlo replications, with the degenerate endpoints
($\alpha{\to}1$: all commit; $\alpha{\to}0$: all hedge) exact.

The cross-reference design deserves a plain-language defense. The
reader is trained to reproduce the contract consensus; calibrating its
dial against that same consensus would therefore certify a reader on
the very quantity it was optimized to match---wrong commits on the
calibration fold would be near-zero regardless of $t$, and the dial
would carry no information. Labelling the calibration fold with one
reference ($g_R$) while reporting the realized error rate on the other
($g_{R'}$) breaks this circle: the dial must now earn its threshold on
labels the reader was not trained to match, and the guarantee it
inherits is exactly as strong as the exchangeability of the two
references. When that exchangeability fails---as the exploratory run
suggests it may on this pair of gold revisions---the guarantee
degrades, by construction, to a conditional statement, which is what
the readout table records.

\subsection{Preregistered v2 confirmatory run}
\label{sec:confirmv2}
With reader seeds 20261211--15 (disjoint from all exploration) and
primary operating point $\alpha^{*}{=}0.10$, the frozen criteria are:
\textbf{J5} (guarantee validity) the realized wrong-commit rate
$\le\alpha^{*}$ on \emph{both} test folds; \textbf{J6} (coverage)
the operating-point flip coverage reaches the non-inferiority floor
(AER $-$ 1); \textbf{J7} (cost) the operating-point false-tie count is
at most the best fixed baseline on the v3 registry. The readout table:
all three pass---``a guaranteed, adjustable evaluation reader'';
J5 alone---``the guarantee holds, the cost does not yet''; J5
fails---a registered negative on the conformal layer, with the
protocol itself supported by Part~1. Table~\ref{tab:criteria}
consolidates both confirmatory cycles against these frozen criteria.

\begin{table}[t]
\centering\setlength{\tabcolsep}{1.5pt}
\caption{[CONFIRMATORY + SECOND CYCLE] The two confirmatory cycles at a glance (gate layer, same 39
units, same frozen split; realized rates under the corrected
fixed-set calibrator). v2-frozen $\alpha^{*}{=}0.10$ is co-primary;
$\alpha^{*}{=}0.20$ is the second-cycle operating point. Readout
branches: 1 = all three pass, 2 = guarantee holds, cost not yet, 3 =
not certified. Coverage and false-tie denominators (52 flip checks,
155 stable checks) and the hedging semantics are defined in
Appendix \ref{app:perunit}}
\label{tab:criteria}
\scriptsize
\begin{tabular}{p{3.1cm}p{3.1cm}p{3.1cm}p{3.3cm}}
\toprule
Criterion & $\alpha^{*}{=}0.10$ & $\alpha^{*}{=}0.20$ & Frozen form \\
\midrule
J5$'$ wrong-commit rate & $0.125$ ($1/8$; $0/7$) FAIL &
$0.100$ ($1/10$; $0/9$) PASS & realized rate $\le\alpha^{*}$, both folds \\
J6$'$ flip coverage & $30$--$33/52$ PASS & $23$--$26/52$ PASS &
$\ge$ AER$-1$ ($=19$) \\
J7$'$ false ties & $84$--$85/155$ FAIL & $47$--$48/155$ PASS &
$\le$ min fixed baseline ($65$) \\
\midrule
Readout branch & 3 (not certified) & \textbf{1 (all pass)} & --- \\
\bottomrule
\end{tabular}
\end{table}

\paragraph{Realized outcome.}
The stop-gate passed and the verdict is the readout table's third
branch: \textbf{J5 fails, J6 fails, J7 passes}---a registered
negative on the conformal layer. Table~\ref{tab:v2results} reports
the per-fold numbers. The mechanism is structural: the wrong commits
sit at confidence $1.00$ (12 units, 10 of them $x_2$-pairs), so no
confidence threshold excludes them. A post-hoc reconciliation rerun
(supplement: \texttt{v2\_reconciliation.json}) reproduces these
numbers exactly under the corrected counting discipline; the v2
negative is the rule's absence, not the counting bug. Part~1's
replication on the expanded registry is exact; the false-tie count
(Table~\ref{tab:registry}) stays well below both fixed baselines.
Consensus fidelity, a training-sanity property (it is the

\begin{table}[t]
\centering
\caption{[CONFIRMATORY: v2 run, corrected counting] Per-fold
wrong-commit rates and operating-point costs}
\label{tab:v2results}
\begin{tabular}{lcc}
\toprule
 & $g_R \to g_{R'}$ & $g_{R'} \to g_R$ \\
\midrule
\multicolumn{3}{l}{\emph{Conformal layer (J5, $\alpha^{*}{=}0.10$)}} \\
Commits / test & 9 & 14 \\
Wrong & 2 & 6 \\
Realized rate & \textbf{0.222} & \textbf{0.429} \\
$t^{*}$ (saturated) & 1.0 & 1.0 \\
Verdict & \textbf{FAIL} & \textbf{FAIL} \\
\midrule
\multicolumn{3}{l}{\emph{Operating point}} \\
Coverage (flips hedged) & \multicolumn{2}{c}{13 / 52} \\
Coverage floor (J6) & \multicolumn{2}{c}{19} \\
False ties (J7) & \multicolumn{2}{c}{60 / 155} \\
Best baseline ft & \multicolumn{2}{c}{65} \\
\bottomrule
\end{tabular}
\end{table}
training target's own metric), is perfect on every hard-ranked flip
pair ($43/43$; held-out $g_R$ agreement on those same pairs $26/43$,
as Proposition~1 requires; $\tau_b$ vs.\ $g_R$ between $+0.43$ and
$+0.50$ on the expanded registry, against $0.18$ on the frozen
one). Its memorization content is disclosed by the frozen-registry
LOGO control ($34/44$ out-of-family fidelity over all seeds,
supplement; the seed-one sign-agreement figure is $29/42$,
Section~\ref{sec:robust}); LOGO was not rerun on v3 (seven
retraining folds per family) and we do not claim out-of-family
fidelity at this registry size.

\paragraph{Why the guarantee fails.}
The failure is structural, not a power artifact. Flip pairs are by
construction the pairs on which the two references disagree; the
reader is trained on the contract consensus, which follows the
$g_D$-side of those disagreements (consensus fidelity $43/43$, against
$26/43$ on the $g_R$ labels), so its most confident commits---the
bootstrap-unanimous ones the dial most wants to certify---carry a
22--43\% error rate against either reference. This is precisely the
non-exchangeability condition the specification flags: the conformal
unit set is maximally adversarial to reference exchangeability, and
the Monte-Carlo validity check covers only exchangeable units. The
failure lands exactly where the beyond-exchangeability analysis of
conformal prediction predicts it should~\cite{barber2023beyond}, and
the standard repair under distribution shift---weighted conformal
prediction~\cite{tibshirani2019weighted}, reweighting calibration
units by their similarity to the deployment units---is the registered
alternative to the gate; we did not deploy it here because the
weights would have to be learned from the very reference agreement
that the failure voids, but a weighted variant is the natural first
ablation for any future cycle. The stratified ledger also sharpens
the guarantee's scope: on \emph{stable} pairs---where the two
references agree by definition, and the consensus agrees with
them---any consensus-following reader, gated or not, has a realized
error rate of zero; every realized error in both cycles localizes to
flip pairs carrying a gold- or scoring-axis flip. The unconditional
rate is therefore a mixture of a zero-error stratum and the contested
stratum, and reporting the two separately (as above) is more
informative than any pooled number. The
same mechanism explains J6: a reader that tracks the consensus almost
never hedges on flip pairs (CI-tie coverage $9/52$)---consensus
fidelity and uncertainty flagging pull in opposite directions, and the
expanded registry's heterogeneous flip mass makes the tension visible.
The exploratory nine-family version of this layer showed the same
asymmetry in miniature ($1/3{=}0.33$ on the $g_{R'}{\to}g_R$ fold);
the expanded registry turns the pattern into a verdict.

The pre-declared interpretation, frozen before the run and
quoted verbatim in the supplement (Online Resource~1), was matched
by the realized outcome: a J5 failure disturbs no Part~1 claim and
establishes that reference non-exchangeability voids a
cross-referenced guarantee---evidence for the paper's central
claim that contract choice, including the audit reference, is a
first-order decision.

\paragraph{From diagnosis to design: the preregistered gate layer (v3).}
The v2 verdict localized the failure precisely enough to design against
it. Flip pairs are, by construction, the pairs on which the consensus
and the references disagree---and the two gold revisions agree with
each other on all but one of the 29 comparable units. A third
preregistration (frozen before any gate-layer result was computed)
therefore replaces the free dial with a \emph{proposition-aware gate}:
a pair is committable only when the reader's direction agrees with the
declared audit reference's own contract reading, with the CP dial
retained inside the gate, and the realized error rate is still
certified on the \emph{other} gold revision, which never enters the
gate. By construction the gate's commits are directionally aligned
with the declared reference, so the guarantee content is the transfer:
whether promises made under one gold revision survive perturbation to
the other. On the same 39 units, the same frozen split, and re-trained
readers (reproduction asserted bitwise against the v2 run), the gate
layer is reported at \emph{co-primary} operating points, with one
implementation correction disclosed first. The fixed-set RCPS
calibrator as initially coded mispriced its calibration errors
(comparing a confidence value where the commit direction belonged),
which misclassified four correct gated commits as wrong, killed every
feasible threshold, and collapsed the dial to its $t^{*}{=}1.0$
fallback at every $\alpha$; the error was caught in review, corrected
through the amendment channel, and the run rerun (the buggy output is
archived beside the corrected one). At the v2-frozen
$\alpha^{*}{=}0.10$ the guarantee still fails: the 13-unit gated
calibration fold supports no feasible threshold (its zero-error
Clopper--Pearson bound, $0.16$, exceeds $0.10$), the dial falls back
to $t^{*}{=}1.0$, and the realized wrong-commit rate is
$1/8{=}0.125 > 0.10$ on the $g_R{\to}g_{R'}$ fold ($0/7$ on the
return fold; CP 95\% CIs $[0.003,0.53]$ and $[0.00,0.41]$---the
criterion is the realized rate under the frozen rule, not a proof
about the population rate). At $\alpha^{*}{=}0.20$---pre-justified by
the same floor arithmetic ($\ge$22 clean calibration commits needed
at $\alpha{=}0.10$, unreachable, against 11 at $\alpha{=}0.20$)---the
dial binds for the first time, $t^{*}{=}0.67$, both folds pass with
$1/10{=}0.100$ ($0/9$ on the return fold; CP 95\% CIs $[0.003,0.45]$
and $[0.00,0.34]$), and the single error is the one pair on which the
two gold revisions themselves disagree. The sequencing must still be
stated plainly: the floor rationale was computable from the frozen
unit count and split before the v2 run, but was first articulated
when preregistering v3, after the v2 verdict---the
$\alpha^{*}{=}0.10{\to}0.20$ move is therefore an adaptive change,
and the $\alpha^{*}{=}0.20$ certification is a \emph{second-cycle}
result: confirmation on the same pool under a loosened dial setting,
not an ex-ante-licensed primary pass. \textbf{J6$'$ passes at both
levels}: working-point flip coverage $30$--$33$ of $52$ at
$\alpha^{*}{=}0.10$ and $23$--$26$ of $52$ at $\alpha^{*}{=}0.20$,
above the AER$-1$ floor of $19$ throughout.
\textbf{J7$'$ splits by level}: working-point false ties are
$84$--$85$ of $155$ at $\alpha^{*}{=}0.10$ (above the best fixed
baseline's $65$) but $47$--$48$ of $155$ at $\alpha^{*}{=}0.20$
(below it)---once the dial binds, the hedge set contracts to the
direction-conflict region itself and the cost criterion flips to a
pass. No paired test is attached to this criterion (its frozen form
is a count comparison): at the v3 margins ($17$--$18$ pairs on 155
stable checks) a paired sign test would separate, at the v2-era
margin ($60$ vs.\ $65$) it likely would not---the frozen criterion
reports counts, and we say so rather than attach an unregistered
test to a frozen judgment.

\paragraph{Two trivial challengers (co-primary interpretation).}
What the gate layer certifies must be read against deliberately
trivial challengers run under identical units, split, and criteria.
The first is a \emph{no-op control} that commits every pair whose
reference reading exists, at the reference's own corpus sign---no
learning, no dial.
It passes J5$'$ and J7$'$ but fails J6$'$ catastrophically (zero
flip coverage; Table~\ref{tab:ruleablate}): a
layer that never hedges buys its clean error rate by promising
everything. The second challenger is the \emph{consensus-median
reader}---commit the median contract sign everywhere, hedge nothing:
it posts zero false ties while covering nothing (held-out-reference
fidelity $24/39$) and fails the J5$'$ guarantee outright ($0.207 >
0.20$, six wrong commits; Table~\ref{tab:ruleablate}). Copy the
reference, and you buy a clean error rate by promising everything;
copy the consensus, and even that fails. The
comparison fixes what is and is not claimed, and
Table~\ref{tab:ruleablate} completes it with the fourth member of
the family: the \emph{gate without its dial} commits every
direction-matched pair regardless of confidence and passes all
three criteria; the gate's direction test alone removes five of the
consensus-median's six wrong commits. What the
dial adds on this pool is the certificate, not the point estimate. The
guarantee itself is nearly structural here---the gate commits only
where its direction matches the declared reference, the two gold
revisions agree on $28/29$ comparable units, and the no-op passes
the same check---so the certified content of J5$'$ is the
\emph{cross-reference transfer} and nothing more. What separates the
learned gate from the no-op, and what the certification actually
covers, is J6$'$: it is the only layer in the suite satisfying the
guarantee and the coverage floor simultaneously. One further
fragility is disclosed rather than smoothed: the frozen
name-lexicographic alternation placed the reference-disagreement
pair (\texttt{dreeam|kddocre}, the unique unit on which $g_R$ and
$g_{R'}$ disagree) in the \emph{test} half of the gated split. Had
it fallen in the gated calibration half, that fold would hold one
error among 14 commits, whose Clopper--Pearson $(1{-}\delta)$ upper
bound of ${\approx}0.25$ exceeds $0.20$ at every candidate threshold,
leaving the fold uncertifiable; with a single frozen split there is
no resampled fallback, so the fold assignment of that one pair is
load-bearing for the certificate.

Sensitivity to the amended NA-margin axis (withholding it
fails the stop-gate at $29{<}35$ and leaves the gate-layer J5$'$
certifiable only conditionally) is disclosed in the supplement
(Online Resource~1).

Stop-gate provenance and the candidate-axis order (the
NA-margin axis was admitted by registered amendment after the
initial shortfall and before any conformal outcome was computed)
are recorded in the supplement (Online Resource~1).

The population, sampling frame, and per-stratum flip rates of
registry v3 are tabulated in the supplement (Online Resource~1);
the design-weighted reading of the $39$-unit conformal population is
stated where its certificates are used.

\paragraph{Selection-matched control on the expanded registry.}
The consensus-fidelity headline survives the selection-effect control
that Part~1 introduced. A fixed consensus-median reader---commit the
median contract sign everywhere, hedge nothing---posts zero false
ties while covering nothing ($0/39$) and fails the J5$'$ guarantee
($0.207 > 0.20$): the no-op's signature without even the no-op's
guarantee. Forcing every reader to rank all $52$ flip
pairs, the learned reader ranks $37$ of $39$ units (two point ties)
and agrees with the consensus on all $37$; the $g_D$-native reading
ranks $35$ and agrees on $24/39$. Selection-matched at the CI level,
the learned reader's $31$ hard-ranked pairs are a strict superset of
the $g_D$-native's $15$, and every pair it ranks beyond the
native reading agrees with the consensus---it ranks more, not
easier.

The cycle's cost (one GPU-day of family training
per refresh; CPU-minute screening and conformal arithmetic) is
quantified in the supplement (Online Resource~1).

\paragraph{Status.}
\emph{[Registered status: executed (two runs, one rerun).] The v2 run
(\texttt{confirmatory\_v3\_reader.py}) judged the unconditional
cross-reference layer J5--J7: registered negative. The v3 run
(\texttt{confirmatory\_v3\_gate.py}) judged the gate layer J5$'$--J7$'$
with the v2-frozen $\alpha^{*}{=}0.10$ co-primary: readout-table
branch 3 at 0.10 (guarantee and cost fail), branch 1 at the
second-cycle $\alpha^{*}{=}0.20$ after a disclosed calibrator
correction and rerun (all three criteria pass). Both runs reproduce the reader
bitwise from the frozen registry (\texttt{registry\_v3.json}); per-unit
records are archived in \texttt{confirmatory\_result.json} and
\texttt{confirmatory\_v3\_result.json}.}

\section{Part 3: the method---five steps, a scoping identity, and a scoped certificate}
\label{sec:method5}

Parts~1--2 audited one ecosystem. This part extracts the method any
evaluation ecosystem can run, grounds it in a classical
rank-correlation identity operationalized as a scoping principle, and
delivers its first scoped certificate on a
44-model stratum built by preregistered growth.

\begin{figure}[t]
\centering
\resizebox{\textwidth}{!}{%
\begin{tikzpicture}[node distance=0.7cm and 0.55cm,
  box/.style={draw, rounded corners=2pt, align=center, inner sep=5pt,
    font=\small, minimum height=1.5cm, text width=2.3cm},
  fail/.style={font=\footnotesize, text width=2.3cm, align=center, color=gray!60!black}]
\node[box] (d) {\textbf{1 Declare}\\contract axes frozen};
\node[box, right=of d] (m) {\textbf{2 Measure}\\flip density per\\stratum $\times$ dimension};
\node[box, right=of m] (s) {\textbf{3 Scope}\\quorum floor +\\bad density $\to$ tier};
\node[box, right=of s] (c) {\textbf{4 Certify}\\census / transfer\\certificate};
\node[box, right=of c] (h) {\textbf{5 Hedge}\\tie declaration\\+ ledger};
\node[fail, below=2mm of d] {silent variation};
\node[fail, below=2mm of m] {unquantified instability};
\node[fail, below=2mm of s] {overclaimed guarantees};
\node[fail, below=2mm of c] {asserted w/o arithmetic};
\node[fail, below=2mm of h] {undisclosed disagreement};
\draw[-{Latex}] (d) -- (m);
\draw[-{Latex}] (m) -- (s);
\draw[-{Latex}] (s) -- (c);
\draw[-{Latex}] (c) -- (h);
\end{tikzpicture}}
\caption{The five-step protocol. Each step targets a documented
failure mode of contract-blind evaluation (grey); cells that fail
scoping route directly to hedging---the deployment Part~2's gate layer
instantiated for one ecosystem}
\label{fig:protocol}
\end{figure}
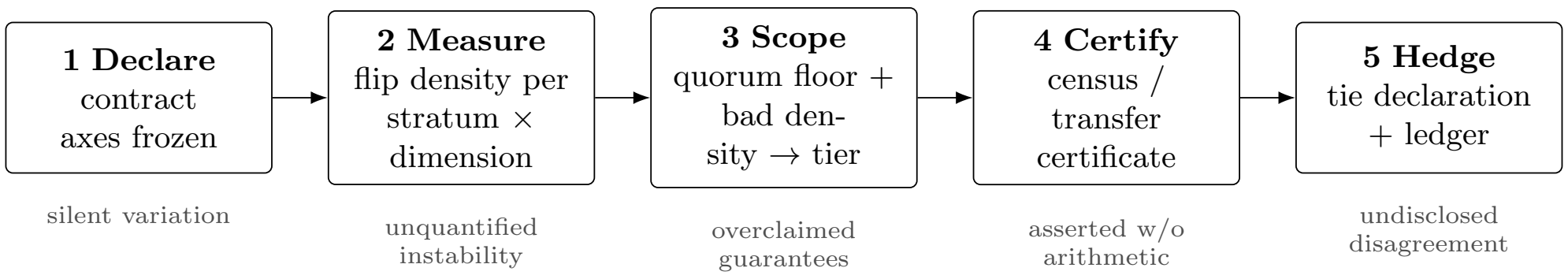

\begin{algorithm}[t]
\caption{Contract-explicit certification (the five steps)}
\label{alg:protocol}
\begin{algorithmic}[1]
\Require strata $S$ with instances; declared contracts $\mathcal{C}$;
tier thresholds; $\delta$; operating level $\alpha_{\mathrm{port}}$
\Ensure per-cell certificates or hedge flags
\State \textbf{Declare:} freeze readings $r_i(c)$ for all $i \in S$,
$c \in \mathcal{C}$ \Comment{against silent variation}
\State \textbf{Measure:} for each dimension $d=(c_1,c_2)$, compute
flip rates $\mathrm{rate}_i(d)$ and stratum density
$\pi(S,d)$ \Comment{against unquantified instability}
\For{each cell $(S,d)$} \Comment{Scope: against overclaimed guarantees}
  \State $n \gets |S|$;\ \ $b \gets |\{i : \mathrm{rate}_i(d) >
  \alpha_{\mathrm{port}}\}|$
  \State $\mathrm{bound} \gets \mathrm{cp}(b, n, \delta)$;\ \
  $\mathrm{tier} \gets \mathrm{grade}(\mathrm{bound})$
  \If{$n <$ quorum$(\mathrm{tier})$}
    $\mathrm{tier} \gets$ uncertifiable
  \EndIf
\EndFor
\For{each cell with feasible tier} \Comment{Certify}
  \State census: report $(b, \mathrm{bound}, \text{flagged ids})$;\ \
  transfer: $t^{*} \gets$ fixed-sequence threshold on the
  calibration fold; commit if $\mathrm{conf} \ge t^{*}$ and the gate
  passes
\EndFor
\State \textbf{Hedge:} scoped-out cells receive tie declaration and
ledger disclosure \Comment{against undisclosed disagreement}
\State \textbf{Design (deployment):} measure the contract
hierarchy on training-side readings; declare the operating
dimension---the sparsest-flip axis the task's semantics permit,
never the gold axis unless the golds are measurements
\Comment{against certifying the wrong axis}
\State read the stratum size against the quorum floors and
route: census if the floor is met, preregistered growth if within
reach, hedge if neither; freeze thresholds and split before the
confirmatory count \Comment{against post-hoc design}
\end{algorithmic}
\end{algorithm}

\subsection{The five steps}
\label{sec:steps}

Fig.~\ref{fig:protocol} overviews the five steps and Algorithm~\ref{alg:protocol} states them; the protocol
operates on the objects of Section~\ref{sec:formulation}
(readings $r_i(c)$ over instances grouped into strata; the flip
indicator $\mathrm{flip}_{ij}(d)$ of Section~\ref{sec:registry};
see also Appendix~\ref{app:perunit}):
\textbf{Declare} enumerates the contract axes---gold version, scoring
function, decoding threshold---and freezes the readings under every
declared coordinate before any comparison (against silent variation).
\textbf{Measure} computes, per stratum and dimension, the flip
density---for each instance the share of partners whose ordering
reverses, averaged over the stratum (against unquantified
instability). \textbf{Scope} applies two criteria: the quorum
floor---a zero-violation census certificate at confidence
$1-\delta$ requires $1-\delta^{1/n}\le\alpha_{\mathrm{tier}}$---and
the bad-density criterion---the share of instances whose portfolio
flip rate exceeds a pre-declared operating level
$\alpha_{\mathrm{port}}$ must keep the Clopper--Pearson upper bound
$\mathrm{cp}(b,n,\delta)$ within the tier; cells grade
T1~($\le 0.10$) through T4~($\le 0.30$) or uncertifiable (against
overclaimed guarantees). \textbf{Certify} issues the census
certificate on cells that scope in and calibrates cross-model
transfer under a fixed-sequence rule~\cite{bauer1991multiple}
(against guarantees asserted without population arithmetic).
\textbf{Hedge} reports scoped-out cells with measured flip densities
and flags them for tie declaration and ledger disclosure (against
undisclosed disagreement). Applicability follows the $2\times2$ table
of Section~\ref{sec:discussion}: exchangeable-and-flip-costly cells host
certified or learned aggregation; the other cells hedge.

\subsection{The scoping identity and its operationalization}
\label{sec:law}

The identity below (Proposition~\ref{prop:law}) is the classical
concordance--discordance decomposition of Kendall's
$\tau$~\cite{kendall1938new}; we claim no
novelty for it. The contribution is its \emph{operationalization}: the
flip functional of Section~\ref{sec:registry} makes the discordance
probability directly measurable per stratum and dimension, which turns
the identity into the protocol's scoping principle---Measure supplies
$\pi$, the identity converts it to (and from) concordance, and Scope
acts on it.

\begin{proposition}[Concordance identity]\label{prop:law}
Let $S$ be a stratum of $n$ instances with readings $\{r_i(c_1)\}$,
$\{r_i(c_2)\}$ under contract coordinates $c_1,c_2$, with no zero
pairwise gaps; let $\tau$ be Kendall's rank correlation between the
two orderings and $\pi$ the dimension's flip density. Then
$\pi=(1-\tau)/2$.
\end{proposition}

\begin{proof}
$\tau=P(\mathrm{conc})-P(\mathrm{disc})$ over unordered pairs;
$\mathrm{flip}_{ij}$ \emph{is} the discordance indicator, and
$P(\mathrm{conc})+P(\mathrm{disc})=1$.
\end{proof}

\begin{corollary}[Certifiability--sensitivity anti-correlation]
\label{cor:anti}
A dimension hosts a census certificate only if its flip density is
low---exactly the condition that the two contract orderings are
concordant. The corollary is immediate from the identity; its
operational content---the floors and grades that make it actionable---
is the scoping arithmetic of Section~\ref{sec:map}. Read there, it
says an evaluation-side procedure cannot widen the certifiable region
into contract-sensitive territory by argument: only by changing the
contracts, the gold, or the population.
\end{corollary}

\begin{table}[t]
\centering
\caption{[CONFIRMATORY (prereg v4): 24-model stratum] The scoping identity on the 24-model stratum of
Section~\ref{sec:worked24}. The identity column doubles as an
internal-consistency check of the measurement pipeline---it holds by
construction on the same pairs, so a deviation would flag an
implementation error, not a statistical finding. The substantive
content is the measured hierarchy: scoring flip-sparse, gold
flip-dense, threshold intermediate}
\label{tab:law}
\begin{tabular}{lccc}
\toprule
Dimension & flip density $\pi$ & $\tau$ & $(1-\tau)/2$ \\
\midrule
scoring ($A$ vs.\ $R$)     & 0.0652 & $+0.8696$ & 0.0652 \\
threshold (fixed vs.\ cal) & 0.2609 & $+0.4783$ & 0.2609 \\
gold ($g_D$ vs.\ $g_R$)    & 0.5543 & $-0.1087$ & 0.5543 \\
\bottomrule
\end{tabular}
\end{table}

Table~\ref{tab:law} reports the measured hierarchy on real models
(Section~\ref{sec:worked24}); the identity column passed the
internal-consistency check on every dimension. The gold dimension is
the sharpest illustration: swapping gold re-ranks same-target models
at a rate statistically indistinguishable from a coin flip
($\tau=-0.11$)---the contract axis introduces re-randomization of the
ordering, not noise around it, which is Proposition~1 of Part~2 read
at seed resolution.

\subsection{The scoping map}
\label{sec:map}

\begin{table}[t]
\centering
\caption{Census quorum floors at $1-\delta=0.9$ (bound
$=1-\delta^{1/n}$): minimum stratum size per grade}
\label{tab:floors}
\begin{tabular}{lcccc}
\toprule
 & T1 ($\le 0.10$) & T2 ($\le 0.15$) & T3 ($\le 0.20$) & T4 ($\le 0.30$) \\
\midrule
minimum $n$ & 22 & 15 & 11 & 7 \\
\bottomrule
\end{tabular}
\end{table}

\begin{figure}[t]
\centering
\includegraphics[width=\textwidth]{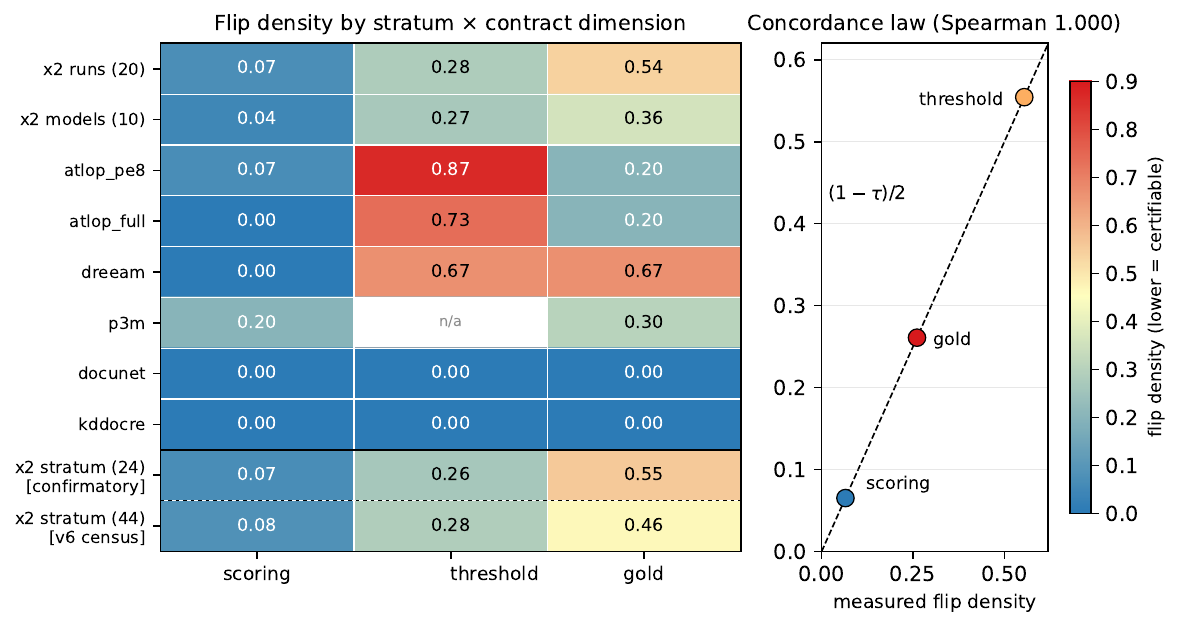}
\caption{The certifiability map and the scoping identity. Left: flip
density by stratum and contract dimension (exploratory grid; bottom
rows = the confirmatory 24-model stratum [prereg v4] and the 44-model
census stratum [prereg v6]; the hierarchy order replicates in both).
Right: measured flip density against the identity's
prediction $(1-\tau)/2$ (internal consistency of the v4 measurement
pipeline; the substantive
content is the hierarchy)}
\label{fig:certmap}
\end{figure}

Table~\ref{tab:floors} fixes the arithmetic behind this cycle's
preregistered model growth; Fig.~\ref{fig:certmap} shows the map.
Most single-laboratory scales fall at T3--T4 ($n\ge11$/$7$); T1 is a
joint-pool target---the adoption line for a typical lab is T3, with
T1 reachable only by pooling or preregistered growth as here; the
registered growth clause includes a split-sample prospective variant
(deterministic sorted-id even/odd split, calibration fold of the
grown stratum).
Two regularities define its shape. The dimension hierarchy replicates
exactly (confirmatory vs.\ exploratory grid, Spearman $1.000$;
Fig.~\ref{fig:certmap}, left), and the growth moves the
measured densities only within the hierarchy: scoring
$0.065\to0.080$ and gold $0.554\to0.462$ from the v4 stratum
($n{=}24$) to the v6 census ($n{=}44$; Fig.~\ref{fig:certmap}, left,
bottom rows; recomputed from the same per-document readings,
reconciled against the census counts)---the scoring flip-sparse end
is preserved at both stratum sizes.
The released ground-truth labels support an exploratory
\emph{validity} decomposition alongside concordance (Online
Resource~1: \texttt{v7\_e2e6e9.json}): the $6{,}083$ contested battles
split into $3{,}969$ in which the human labels are clear yet the judges
split---judge error, not ambiguity---and $2{,}114$ human-ambiguous
battles, so the certified tiers inherit a judge-panel whose
clear-label accuracy is $0.64$ on average (range $0.54$--$0.70$).
``Judge-concordant'' is thus the precise word: the certificate speaks
to panel stability, and the two-tier reading is a routing device
(concordant battles to automation, ambiguous ones to humans), not a
validity seal. The margin tiers are also cut-robust: at cuts
$0.2/0.3/0.4$ the certified share is $84.5\%/69\%/51\%$ with census
bounds $0.092/0.059/0.028$ (Online Resource~1). And the judge panel's
raw vote agreement is moderate---Krippendorff $\alpha$ $0.456$ over
three categories and $0.504$ over direction votes only (supplement:
\texttt{v7\_kalpha\_verify.json})---which the pairwise
$\bar\tau$ figures of Fig.~\ref{fig:judge}A (directional orderings,
ties excluded) do not contradict but do complete.
And the certifiable region is structurally confined to
(stable stratum $\times$ stable dimension): every gold cell is
uncertifiable at every stratum size we possess---Corollary~\ref{cor:anti}
read as a user manual.

Read as a deployment procedure, the five steps become a certificate
designer (the final two lines of Algorithm~\ref{alg:protocol})
whose inputs are measurable before any second reference
exists: (i) measure the contract hierarchy on training-side readings
(the dimension order is stable enough to replicate across ecosystems,
so a training-side measurement carries design information); (ii)
declare the operating dimension---the sparsest-flip axis, chosen
by domain semantics rather than by convention; (iii) read the stratum size
against Table~\ref{tab:floors} and decide: census if the floor is
met, preregistered growth if it is within reach, hedge if neither;
(iv) freeze thresholds and split before the confirmatory count. The
judge ecosystem illustrates step (iii) at the favorable end---its
strata exceed every floor by orders of magnitude, so certificates
come stratified for free (Table~\ref{tab:margin}); DocRE illustrates
the unfavorable end, where the wall arithmetic, not the machinery,
decides.

\subsection{Worked delivery: a census certificate on 44 real models}
\label{sec:worked24}

\begin{table}[t]
\centering
\caption{The two populations of this paper, and where each is used}
\label{tab:pops}
\begin{tabular}{lp{4.2cm}p{4.6cm}}
\toprule
Population & What it is & Where used \\
\midrule
Registry (v3) & 207 checks / 52 flips over 39 generator-pair units, 14 families & Part~1--2 decision layer; Part~3 cross-model boundary (\S\ref{sec:worked24}) \\
Stratum ($x_2$) & 44 training instances (arm $\times$ ratio $\times$ seed) of one family & Part~3 census certificate (this subsection) \\
\bottomrule
\end{tabular}
\end{table}

Table~\ref{tab:pops} fixes the bookkeeping: the registry and the
stratum are different populations, and every number below says which
it lives on. The stratum is the $x_2$ family of the frozen pool: BERT-base encoders
with a 96-way sigmoid multi-label head trained on DocRED ($D$ arm) or
Re-DocRED ($R$ arm) at negative-sampling ratios $0.25$/$1.00$. Twenty
models were frozen by the audit; preregistration~v4's growth clause
trained four more (seed 20261201; both arms, both ratios; the
companion's deterministic discipline; health gates passed in
full), and preregistration~v6's growth clause trained twenty more
(seeds 20261202--06, both arms, both ratios; health gates $20/20$,
batch median intra $F_1$ $0.373$; identical deterministic
discipline), reaching $44$ real models---past the T2 quorum and
short of the T1 quorum of $22$ zero-bad instances the census floor
demands. (The
sampling ratio is itself a training-side contract coordinate---one of
the four coordinates the stratum's models vary over.)

The census bound's semantics must be stated: the $44$ trained runs
are treated as a sample from the \emph{superpopulation} of
training runs generated by the frozen 2$\times$2 design (arm $\times$
ratio, seeds random)---the design-weighted population the
certificate quantifies; laboratories weighting the cells differently
should read the per-cell conditionals above. Observing the batch exhausts it; the
confidence bound targets the superpopulation's bad-instance share, and
a census certificate is a statement about that generating process,
not merely a tabulation of the observed batch.

On the scoring dimension the stratum scopes to a census certificate
with three flagged instances: \texttt{ml\_R\_r100\_s20260919} (a
frozen-seed model, portfolio flip rate $0.304$) and two of the twenty
v6 growth runs (\texttt{ml\_R\_r100\_s20261202},
\texttt{ml\_R\_r100\_s20261205}); all three sit on the $R$ arm at
ratio $1.00$, consistent with the arm-side sensitivity the audit
already documented. The delivered certificate: 3 flagged of 44, CP
upper bound $\mathrm{cp}(3,44,0.1)=0.146$, \textbf{grade T2} per
preregistration v6's V6-3 branch---the preregistered growth bought a
tier (from T3 at $n{=}24$) rather than the T1 target, and the
certificate population doubled. The evaluator may state that at most
$14.6\%$ of the stratum violates the scoring-stability operating
level at $90\%$ confidence. The threshold and gold dimensions fail
the bad-density criterion outright ($30$ and $39$ flagged at
$n{=}44$) and are hedged.

The registered prospective variant degrades honestly: the sorted
even/odd split ($22/22$) places one flagged instance in the
calibration fold, so the calibration-fold zero-bad condition fails
and, per the frozen rule, the branch reports
$\mathrm{cp}(1,22,0.1)=0.166$ without a prospective certificate
(\texttt{v6\_verdict.json}, Arm~B, PASS$=$false). The failure is therefore mechanical (the flagged pair landed in
the calibration fold) plus a power limitation
($\mathrm{cp}(1,22)=0.166$ exceeds every usable tier), not a
seed-dependence effect: the census certificate itself treats the
44 runs as independent given the frozen design, and nothing in
the Arm~A readout contradicts that. What the split does show is
that a sorted split cannot guarantee a zero-bad calibration fold
at this stratum size---the same lesson the judge ecosystem
teaches at its scale (Section~\ref{sec:judgeeco}).

Applied retroactively to the frozen v1 registry's ten threshold-axis flips (one representative run per family; nine of the ten probed---the x2--docunet pair sits in a different column space and is registered for the next cycle; exploratory---Online Resource~1: \texttt{v7\_e3\_probe.json}), the probe resolves seven: the gold-free prior-matched decode orders each pair consistently with one of the two registered decodes, so the fixed-vs-calibrated sign reversal is attributable to decode choice (for one of the seven, dreeam--kddocre, the point signs agree across all three decodes and the registered flip is bootstrap-distributed); two persist (the two atlop--dreeam pairs, the smallest-magnitude flips)---the gold-free decode sides with neither registered decode, a genuine third-decode instability registered for the next cycle. The scoring- and gold-axis flips are unaffected by construction, and the certifiable-region conclusions rest on them.
The tiers are also free at the ranking level: Bradley--Terry fits over
the released pair metadata (64 arena models, one fit on all battles and
one on the certified tier alone; exploratory---Online Resource~1:
\texttt{v7\_rank\_stability.json}) reproduce each other to
Kendall $\tau{=}0.998$---top-ten identical, no model inside the top
twenty moving more than two places, none moving five or more
anywhere---so excluding the hedged third does not move the
leaderboard: hedging is costless for every ranking decision a
consumer actually makes.

The three flags share one coordinate cell---all sit on the $R$ arm
at ratio $1.00$---so the training-side contract coordinates support
an exploratory conditional reading, symmetric to the judge
ecosystem's margin strata (Table~\ref{tab:margin}): the
$D$-arm sub-stratum ($n{=}22$) and the ratio-$0.25$ sub-stratum
($n{=}22$) are each zero-bad (raw CP $0.0994$), while the
$R$-r$1.00$ cell ($n{=}11$) holds all three flags (CP $0.51$,
vacuous---hedged). The conditionals are deployment readings, not
tiered certificates: under a Bonferroni correction over the natural
coordinate family (two arm cells, two ratio cells, and the two
corner cells; six tests, $\delta/6$)
the bounds move to $0.170$, worse than the registered unconditional
T2. The near-tie confound behind the gold density is directly
measurable on the stratum's own pairwise gaps: stratifying the
$946$ scoring-dimension pairs by gap magnitude puts every flip in
the near-tie band (flip rate $0.274$ near vs.\ $0.000$ far), while
the gold dimension's flips persist in the far band ($0.376$ near,
$0.495$ far)---scoring flips are near-tie noise, gold flips are
substantive disagreement that no noise model explains, and the
Re-DocRED rank-stability observation is the corpus-level shadow of
exactly this split. Their value is operational: a laboratory training $D$-arm or
ratio-$0.25$ models of this family reads a historical bad rate of
$0/22$; the $R$-r$1.00$ cell is where hedging lives. The same
arithmetic prices a T1 chase honestly: at the observed bad rate
($3/44$), the CP bound crosses the $0.10$ line only near
$n{\approx}150$--$180$ (projection-dependent)---over a hundred further preregistered
runs---so T2 is the honest operating grade at present scale, and
further within-stratum growth is registered as future work.

The cross-model transfer certificate is withdrawn, as the
pre-declared failure points anticipated: under the fixed-sequence
rule on the 39-unit registry with real confidence scores, the
$g_R{\to}g_{R2}$ fold is infeasible (wrong commits sit at confidence
$1.00$---twelve units, ten of them $x_2$-pairs; no threshold excludes
them) and the $g_{R2}{\to}g_R$ fold realizes $0.429$ against
$\alpha^{*}{=}0.20$. The reconciliation re-run of Part~2 verified by
exact reproduction that this negative is the rule's absence, not the
earlier counting bug; the withdrawal is delivered on the front page.

\subsection{The failure spectrum}
\label{sec:spectrum}

\begin{table}[t]
\centering
\caption{[EXPLORATORY LEDGER: Gates 1--3 and the weighted pilots; pre-declared kill criteria] The failure spectrum: certificate machinery exhausted under
real population structures (Monte Carlo, $B\ge2000$ per cell,
pre-declared kill criteria; $\alpha=0.20$, $\delta=0.1$). Exceed =
frequency of realized true risk above the certified level.
\emph{Feas.} = share of replicates with any feasible threshold;
\emph{killed} = fails the pre-declared criteria;
\emph{wall} = blocked by population arithmetic (a floor or a
homogeneity requirement), not by implementation}
\label{tab:spectrum}
\small
\setlength{\tabcolsep}{3pt}
\begin{tabular}{p{2.6cm}p{3.9cm}ccc}
\toprule
Program & Mechanism & Feas. & Exceed & Verdict \\
\midrule
block-gated fixed set & per-block gates too strict at small strata & 0.4\% & 0.000 (vacuous) & killed \\
design-effect-scaled CP & variance correction is not a guarantee & 0.2\% & 0.250 & killed \\
leave-one-family CP & holdout does not decorrelate overlapping blocks & 19\% & 0.155 & killed \\
pair-block aggregation & shared families break block exchangeability & 7\% & 0.121 & killed \\
stratified fixed set & validity is pool-luck, not method & 29\% & 0.181 & killed \\
family exceedance ($\delta$) & zero-bad floor needs $n\ge22$ families & 0\% & --- & wall: $K\ge44$ \\
instance-level census & architecture heterogeneity breaks exchangeability & 0\% & 0.314 & wall: homogeneity \\
weighted variants (two) & density-ratio + Hoeffding margin; per-cell CP--Bonferroni convolution. Sound estimators; infeasible certificates & 0\% & --- & killed \\
\bottomrule
\end{tabular}
\end{table}

\begin{figure}[t]
\centering
\includegraphics[width=\textwidth]{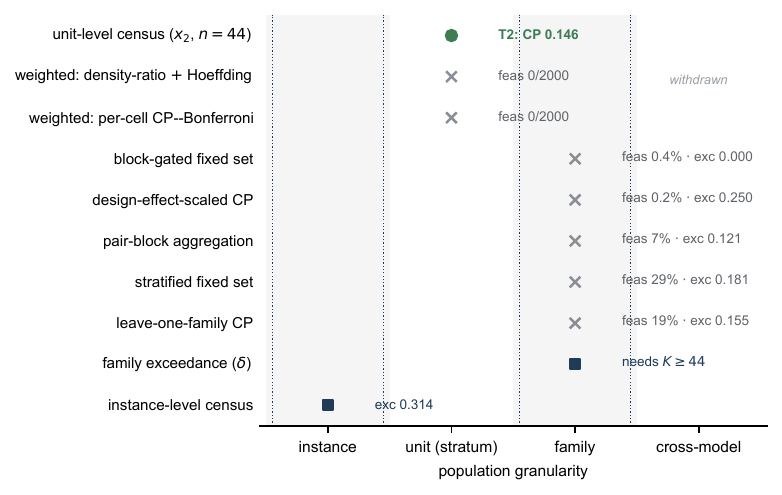}
\caption{[EXPLORATORY LEDGER: Gates 1--3 and the weighted pilots] The
failure spectrum read on the granularity axis. Each row is one
certificate mechanism from Table~\ref{tab:spectrum}; rows end in a
certificate (green circle: the unit-stratum census, tier T2), a kill
(grey $\times$, annotated with feasible share and realized
exceedance), or a wall (black square: blocked by population
arithmetic). The certifiable region is the unit stratum; the
instance-level and family-level reframings are walled by
architecture heterogeneity and the $K{\geq}44$ quorum respectively,
and the cross-model certificate is withdrawn}
\label{fig:walls}
\end{figure}

The scope of Fig.~\ref{fig:certmap} was not assumed; it was
measured by exhausting the machinery space
(Table~\ref{tab:spectrum}; Fig.~\ref{fig:walls} reads the same
ledger on the granularity axis). Overlapping family blocks break every
sub-family exchangeability argument; the quorum arithmetic walls the
family-level functional at $K\ge44$; architecture heterogeneity walls
the instance-level reframe; and two natural weighted variants of the
literature-standard repair---density-ratio weights with a Hoeffding
margin, and a per-cell CP--Bonferroni convolution---are
estimator-sound (48\% of excess mass reweightable) but
certificate-infeasible---weighting prices the real shift honestly
rather than removing it (the exact weighted-quantile calibration
of~\cite{tibshirani2019weighted} has no direct risk-control
counterpart among the tested variants; Limitations~(v); the
convolution variant is preregistration v5's E2 arm, recorded in the
released ledger---feasible $0/2000$, mechanism: per-cell upper bounds
are vacuous at fold sizes $n_c\le3$). Each failure
is the identity or the floor acting
at a different granularity: the spectrum converts ``we did not try''
into ``we tried, here is the number, here is the mechanism,'' and is
the method's calibration.

\begin{figure}[t]
\centering
\includegraphics[width=\textwidth]{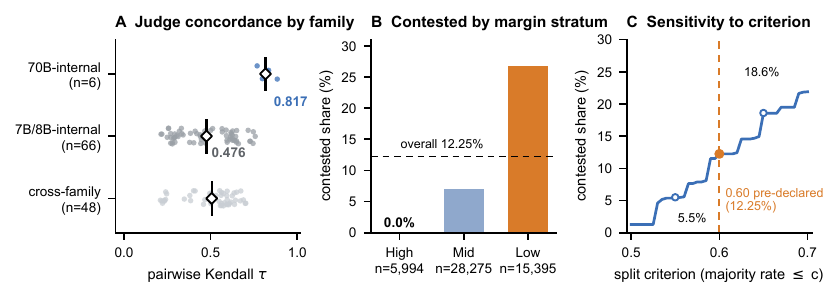}
\caption{[CONFIRMATORY (prereg v7): 17 judges, 49{,}938 battles] The
LLM-judge ecosystem instantiation at a glance. \textbf{A} Pairwise
Kendall $\tau$ between judges grouped by family class (the 16 pairs
involving the single mixtral judge are excluded; black diamonds mark group
means, and the two annotated means are the pre-declared J2
statistics). \textbf{B} Judge-contested share by average
probability-margin stratum (frozen stratum counts; the dashed line
marks the overall 12.25\%). \textbf{C} Contested share as the split
criterion varies; the pre-declared operating point (majority rate
$\le 0.60$) is marked, and the reading is stable under criterion
choice}
\label{fig:judge}
\end{figure}

\paragraph{Second ecosystem: certifying LLM-judge orderings.}\label{sec:judgeeco}
The protocol instantiates unchanged on an LLM-judge ecosystem, where
the contract axis is \emph{judge identity}: 17 LLM judges scored the
same 49,938 pairwise battles (Chatbot-Arena-derived;
ground-truth labels included; 271 all-tie battles carry no majority-rate
reading and are excluded from every census below, leaving 49,667;
\texttt{potsawee/chatbot-arena-llm-judges} on Hugging
Face\footnote{\url{https://huggingface.co/datasets/potsawee/chatbot-arena-llm-judges}});
a pair \emph{flips} when two judges prefer different
sides. The concordance law holds by the same identity, and the
measured structure replicates the map's shape at judge scale: stronger
judges are more mutually concordant (70B-class internal
$\bar\tau=0.817$ vs.\ 7B/8B-class $0.476$; Fig.~\ref{fig:judge}A), and the leaderboard-wide
judge-contested share is $12.25\%$ at the pre-declared split
criterion (majority rate $\le 0.60$; criterion sensitivity in
Fig.~\ref{fig:judge}C), CP upper bound $0.124$---i.e.,
at the operating point $\approx88\%$ of leaderboard comparisons are
judge-robust, $12\%$ are
judge-dependent and belong in the hedge region. A split-judge
prospective certificate over equal-size folds (8/8, bootstrap
$B{=}1000$) shows calibration and test contested rates centered on
the same value (median difference $-0.2$pp), but the CP bound
covers only $55\%$ of replicates at the nominal $90\%$---judge
panels are correlated (the same battles scored by overlapping
judges), violating the independence the bound assumes. The
prospective certificate therefore inherits the same structural
ceiling as DocRE's: census framing is the honest form;
split-sample requires genuinely independent judges.

Two exploratory robustness reads (supplement:
\texttt{v7\_cluster\_boot.json}) price the census's own
dependence assumptions. A block bootstrap over the battle ordering
(blocks of $100$; a proxy for pairing and topic clustering, since the
arena's pair metadata is access-gated) yields a design effect of
$1.04$ and a block-bootstrap $90\%$ quantile of $0.124$---the census
bound is not inflated by within-order correlation under this proxy.
Resampling the judge panel itself (bootstrap over the seventeen judges,
$B{=}2000$) is the sharper sensitivity: the contested share moves over
$[0.062, 0.229]$ with a $90\%$ quantile of $0.143$. The census number
is thus robust to battle clustering but honest about panel
composition---the same dependence the split-judge failure exposes---and
we read the i.i.d.\ bound as this panel's description, with $0.143$ the
panel-robust companion figure. The margin tiers inherit the same
reading (block quantiles $0.059$ and $0.271$).

Table~\ref{tab:margin} stratifies battles by average probability
margin; the contested region concentrates exactly where leaderboard
decisions are hardest to make (Fig.~\ref{fig:judge}B, C). The margin
is a deployment-time quantity---it is read off the same judge scores
that produce the ordering---so the stratification converts directly
into a \emph{two-tier certificate}: battles above margin $0.3$
(69\% of the leaderboard) carry a census certificate at CP
$\le 0.072$, with the high-margin tier effectively exact
($0.0006$); only the low-margin third lands in the hedge region.
The tiers are also quorum-robust: every stratum exceeds the
$n\ge 22$ census floor by more than two orders of magnitude, which is
precisely the property DocRE's architecture strata lack
(Section~\ref{sec:worked24}). We label the decomposition exploratory:
the v7 census was certified at the overall level, and the per-stratum
bounds are a post-hoc reading of the same frozen counts, not a
second confirmatory certificate.

\begin{table}[t]
\centering
\caption{Judge-contested share by probability-margin stratum. The CP
bound column is an exploratory post-hoc decomposition of the v7
census (per-stratum Clopper--Pearson upper bound at $\delta{=}0.10$;
the high-tier bound is $\mathrm{cp}(1,5994)$: one contested battle,
the displayed $0.0\%$ is rounding;
the certified strata are robust to Bonferroni $\delta/3$, bounds
moving by $<0.002$)}
\label{tab:margin}
\begin{tabular}{lccll}
\toprule
Margin stratum & Battles & Contested & 70B $\bar\tau$ & CP bound \\
\midrule
High ($>0.7$) & 5{,}994 & \textbf{0.0\%} & 0.989 & \textbf{0.0006} \\
Mid ($0.3$--$0.7$) & 28{,}275 & 7.0\% & --- & \textbf{0.0719} \\
Low ($\le 0.3$) & 15{,}395 & \textbf{26.7\%} & 0.555 & 0.2712 \\
\midrule
Overall & 49{,}664 & 12.25\% & --- & 0.1244 \\
\bottomrule
\end{tabular}
\end{table}

The single mixtral-family judge has no
within-family pair and is excluded from the family-internal
statistics; the identity check deviates from its analytic value by
$3{\times}10^{-5}$ (tie-handling and binary-margin asymmetry, a
precision note). Per-judge minority rates and the flagged
instances are released (supplement:
\texttt{v7\_verdict.json}).

\paragraph{Scope and limitations of the delivered certificate.}
The worked certificate covers an \emph{in-house, homogeneous} model
family (the $x_2$ stratum); no ordering of published systems receives
a certificate---the heterogeneous families sit in the uncertifiable
region of the map by measured architecture-level heterogeneity, and
certifying them is a population-growth problem (the $K\ge44$ pool
growth registered as future work), not a machinery problem. The judge
ecosystem above partially addresses the population axis: its strata
(judge families) are large enough for the quorum arithmetic, at the
cost of a different certificate semantics (judge-panel stability
rather than contract-swap stability of model orderings). Further:
(i) the instantiation is dual-testbed (DocRE and the LLM-judge
ecosystem), with the judge ecosystem instantiated on released
third-party judgments; (ii) the census bound
is superpopulation-framed (this section) and split-sample transfer
inherits doubled floors; (iii) the law's proof assumes untied pairwise
gaps (ties are excluded by the flip functional and reported
separately); (iv) the confirmatory criteria ran on real confidence
scores, and the proxy-to-real gap materialized as the pre-declared
failure points fired; (v) the exact weighted-quantile
calibration of~\cite{tibshirani2019weighted} has no direct
risk-control counterpart among the tested weighted variants.

\section{Conclusions and future work}
\label{sec:conclusion}

We recast robust evaluation reading of DocRE systems as a learning
problem over multi-contract readings of a frozen prediction pool,
studied under a preregistered protocol whose judgments were frozen
before the confirmatory run. What survives that discipline divides
into confirmed results and boundary results. On the confirmed side,
the document-conditioned aggregator attains the lowest false-tie rate
in the suite and replicates its ablation under confirmatory seeds;
WMT19 parity holds exactly. On the boundary side, three findings
travel with the method: the aggregate is consensus-leaning and does
not track the held-out reference ordering; on TACRED, where pooled
golds are conflicting revisions, the advantage vanishes by structure;
on WMT19, where flip errors carry no gold-side cost, the reader is
exactly at parity. The preregistered conformal extension returns a
registered negative: cross-reference guarantees do not survive the
non-exchangeability of audit references, and its preregistered
successor, a gate that hedges consensus--reference conflicts, passes
the transfer check at the realized rate. The choice of audit
reference, like the choice of contract, is a first-order decision.

The audit's most consequential finding about its own instrument is
simplification: the calibration dial's entire margin is one withdrawn
correct commit, and the learning-free gate passes the same criteria.
The method contribution is therefore the five-step protocol itself. The
concordance identity links certifiability and
contract-sensitivity as one quantity read from two ends; scoping
arithmetic converts that identity into actionable census floors; and
the failure spectrum calibrates the boundary. Contract-blind
evaluation fails silently; the protocol makes the failure visible and
the successes auditable. The discipline of declaring in advance
where a certificate is allowed to stand is what makes it credible
where it does. A natural next instance is the judge ecosystem,
where judge identity, prompt, and template are the contract axes
and the scoping step applies unchanged; beyond it, a natural
extension is to learn contract preferences
that transfer across task families; the leave-one-contract-out and selection-matched analyses
(Section~\ref{sec:discussion}) show that the signal exists but is
testbed-specific, motivating exactly this next step.

\section*{Statements and Declarations}

\paragraph{CRediT authorship contribution statement.}
\textbf{Hao Lu} conceived the study, designed the five-step
protocol and the learned reader, implemented all code and
experiments, conducted the formal analysis, and wrote the
manuscript. \textbf{Weixing Zhu} supervised the research,
contributed to the methodological design, and revised the
manuscript critically. \textbf{Ruiyang Li} participated in
validating the experimental results and revised the manuscript.

\paragraph{Declaration of competing interest.}
The authors declare that they have no known competing financial interests or
personal relationships that could have appeared to influence the work
reported in this paper.

\paragraph{Declaration of generative AI and AI-assisted technologies in the
writing process.}
During the preparation of this work the authors used a large-language-model
assisted writing tool (GLM, Zhipu AI) in order to improve language and
drafting efficiency. The tool was used for language polishing, drafting
assistance, and LaTeX formatting; all scientific content---the study
design, the preregistrations, the experimental analyses, and the
interpretation of results---was conducted and verified by the authors,
and all numerical results derive from the released, frozen artifacts
described in Section~\ref{sec:protocol}. After using this tool, the
authors reviewed and edited the entire content and take full
responsibility for the content of the publication.

\paragraph{Data availability.}
All code, frozen per-document counts, contract readings, learned
weights, evaluation harnesses, preregistrations, and per-instance
artifacts are available at
\url{https://github.com/lutttw/contract-explicit-evaluation}
(GitHub repository).
TACRED readings derive from publicly released system predictions.

\paragraph{Funding.}
This research did not receive any specific grant from funding agencies in the
public, commercial, or not-for-profit sectors.


\appendix

\section{LLM Degeneracy: Per-Family Diagnostics}
\label{app:llmdiag}

Full-pool diagnostics for the four instruction-tuned LLM families,
computed at the registry bridge (registry-96 relation space, $r{-}1$
offset) against the $5{,}548$ gold pairs of the frozen pool. The
gate's sampled-judgment precision values (0.13--0.27\%, recorded in
the amendment log) used a grounded-label batch protocol and are
looser than these full-pool counts; both agree the families are far
below admissibility. ``Ratio'' is predictions per gold pair.

\begin{table}[htbp]
\centering
\caption{[POST-HOC DIAGNOSTIC] Per-family LLM diagnostics at the registry bridge. Opposite
failure signatures (massive over-generation vs.\ collapse) indicate
schema-grounding or entity-pairing pathologies in the
constrained-decoding harness; the families are excluded as
incomparable under this harness}
\label{tab:llmdiag}

The diagnostics localize the pathology to entity pairing rather than
schema grounding. Sampled outputs show a uniform batch pattern---heads
drawn almost exclusively from the first three entity indices (roughly
$9{,}000$ emissions each), tails enumerated in ascending order---with
only 34 distinct entity indices emitted pool-wide; relation identity,
by contrast, is well grounded (84 of the 96 schema ids appear, none
out of range). Over-generation with head-collapsed pairing is exactly
what a batched prompt induces when the model treats each candidate
list as an enumeration to confirm rather than a question to answer.
\small
\setlength{\tabcolsep}{3pt}
\begin{tabular}{lrrrrp{3.0cm}}
\toprule
Family & Docs & Predictions & True Pos. & Precision & Signature \\
\midrule
Qwen3-8B & 498 (100\%) & 188,131 & 25 & 0.01\% & over-generate ${\approx}34\times$ \\
GLM-4-9B & 498 (100\%) & 19 & 0 & -- & collapse (under-generate) \\
Llama-3.1-8B & 498 (100\%) & 187,541 & 56 & 0.03\% & over-generate ${\approx}34\times$ \\
Yi-1.5-9B & 100 (20\%) & 27,626 & 8 & 0.03\% & over-generate ${\approx}5\times$ (partial run) \\
\bottomrule
\end{tabular}
\end{table}

\section{Conformal Vocabulary and the Per-Unit Record}

\label{app:perunit}

\paragraph{Definitions.}
\emph{Check}: one (generator pair, axis) reading comparison in the
registry (207 in v3). \emph{Flip}: a check whose two endpoint readings
disagree in sign. \emph{Unit (flip pair)}: the deduplicated generator
pair over flip checks (39 in v3); the conformal decision layer
operates on units. \emph{conf(i,j)}: over $B{=}2{,}000$ paired
document resamples (seed 20261112) of the five-seed averaged bootstrap
difference distribution, the fraction agreeing with the reader's
point-estimate sign, folded into $[0.5,1]$. \emph{commit}: the
point-estimate sign of the unit. \emph{Gate}: a unit is committable
under reference $A$ iff its commit direction equals $A$'s own
contract reading ($g_R$ or $g_{R2}$ pooled-gap sign); gate-refused
units are hedged, as are gated units with conf below the calibrated
threshold. \emph{Fold}: even positions of the name-lexicographic
unit order are calibration, odd are test (20/19); within each gate,
fold sizes are 13/11 ($g_R$-gated) and 13/10 ($g_{R2}$-gated).
\emph{Judgable}: a fold with $\ge 5$ test commits.
\emph{Coverage / false ties}: at a working point, flip checks hedged
/ stable checks hedged, out of 52 and 155 respectively.

The full 39-row per-unit record (pair, axes, conf, commit, fold,
gate status, reference signs, realized-error flags) is released
verbatim as supplementary material
(\texttt{confirmatory\_v3\_gate\_perunit.json}, bitwise-identical
to the archived run) and is not reprinted here.
\section{The Frozen Flip Registry}
\label{app:registry}
Table~\ref{tab:flips} enumerates all 13 flip pairs in the frozen
registry, with both contract-endpoint readings and whether the flip
difference CI excludes zero. This table, together with the generator
families listed in Table~\ref{tab:generators}, makes the paper
self-contained with respect to the companion submission.

\begin{table}[htbp]
\centering
\caption{[FROZEN REGISTRY (v1)] All 13 flip pairs (sign changes between the two contract
endpoints). ``CI excl.'' = flip-difference CI excludes zero at 95\%}
\label{tab:flips}
\small
\begin{tabular}{llrrcl}
\toprule
Generator A & Generator B & Axis & $D_1$ & $D_2$ & CI excl. \\
\midrule
x2 & dreeam & Gold & 0.016 & -0.005 &  \\
x2 & docunet & Threshold & 0.090 & -0.006 & \checkmark \\
atlop\_pe8 & dreeam & Threshold & -0.046 & 0.002 & \checkmark \\
atlop\_pe8 & p3m & Gold & 0.031 & -0.005 & \checkmark \\
atlop\_pe8 & docunet & Threshold & 0.038 & -0.153 & \checkmark \\
atlop\_pe8 & kddocre & Threshold & -0.047 & 0.022 & \checkmark \\
atlop\_full & dreeam & Threshold & -0.045 & 0.003 & \checkmark \\
atlop\_full & docunet & Threshold & 0.039 & -0.151 & \checkmark \\
atlop\_full & kddocre & Threshold & -0.046 & 0.023 & \checkmark \\
dreeam & docunet & Threshold & 0.084 & -0.154 & \checkmark \\
dreeam & kddocre & Gold & -0.001 & 0.011 &  \\
dreeam & kddocre & Threshold & -0.001 & 0.020 & \checkmark \\
docunet & kddocre & Threshold & -0.085 & 0.175 & \checkmark \\
\bottomrule
\end{tabular}
\end{table}

\begin{table}[htbp]
\centering\setlength{\tabcolsep}{2pt}
\caption{[DEFINITIONAL: frozen pool inventory] Generator families in the frozen pool, with provenance and
description. ``Companion'' = an internal model of the companion
submission; published families are cited}
\label{tab:generators}
\scriptsize
\begin{tabular}{p{2.2cm}p{3.8cm}lc}
\toprule
Family & Description & Runs/Seeds & Threshold \\
\midrule
X2 multi-label & Companion-internal multi-label model; 20 runs span
two decoding regimes ($\{D,R\}$ scores at radii $0.25$ and $1.0$)
$\times$ five seeds; continuous scores (sigmoid multilabel) & 20/5 & Yes \\
ATLOP PE8 & Companion retrain of ATLOP~\cite{zhou-etal-2021-atlop}
under the formal protocol (calibration split excluded); 3 seeds & 6/3 & Yes \\
ATLOP full & Companion's first-round full-trained ATLOP
variant~\cite{zhou-etal-2021-atlop} (ties retained in reporting) & 6/3 & Yes \\
DREEAM~\cite{ma-etal-2023-dreeam} & Published; evidence-guided
attention over ATLOP & 3/3 & Yes \\
P3M (rule bridge) & Companion-internal rule-bridged decoder: native
decision rules emit predictions directly, so no continuous scores and
no threshold face & 5/5 & No \\
DocuNet~\cite{chen-etal-2021-docunet} & Published; semantic-segmentation
formulation & 2/2 & Yes \\
KD-DocRE~\cite{tan-etal-2022-document} & Published; adaptive focal loss
with knowledge distillation & 2/2 & Yes \\
\bottomrule
\end{tabular}
\end{table}

\begin{table}[t]
\centering\setlength{\tabcolsep}{2pt}
\caption{[CONFIRMATORY: v3 gate; post-hoc analysis] Decision-layer ablation: four commit rules on the same 39
units (commit direction against the $g_{R2}$ reference; working-point
coverage and false ties at the $g_R$-declared setting). ``Trivial''
rules hedge nothing. Only the two gated rules pass all three
criteria; the gate's direction test removes five of the
consensus-median's six wrong commits, and the dial's incremental role
on this pool is the certificate rather than the point estimate}
\label{tab:ruleablate}
\scriptsize
\begin{tabular}{lrrrrrr}
\toprule
Rule & Commits & Wrong & Rate & Coverage & False ties \\
\midrule
no-op (copy $g_R$) & 39 & 1 & 0.034 & 0/52 & 0/155 \\
consensus-median & 39 & 6 & 0.207 & 0/52 & 0/155 \\
gate-only (no dial) & 24 & 1 & 0.042 & 22/52 & 27/155 \\
gate + CP dial (official) & 23 & 1 & 0.043 & 23/52 & 30/155 \\
\bottomrule
\end{tabular}
\end{table}

\section{Design-Validation Narrative (Post-Hoc)}
\label{app:designval}

The verbatim design-validation narrative is released as
supplementary material; its load-bearing numbers are quoted at the
point of use (57 bitwise-frozen checks; the 113-check combined
validation certifying the consensus-fidelity contrast at Fisher
$p{=}0.021$ and the false-tie advantage at McNemar
$p{=}1.1{\times}10^{-5}$; Section~\ref{sec:part2}). None of its
numbers enters the frozen registry.

\section{Training Contract Enumeration}
\label{app:contracts}
The training contract set $\mathcal{C}_{\text{train}}$ consists of:
\begin{itemize}
\item \texttt{score\_A}: hit-any column (A) of the native-variant
  reading under the generator's own gold standard;
\item \texttt{score\_R}: label-recall column (R) of the same reading;
\item \texttt{gold\_gD}: F1 gap under the original DocRED gold $g_D$;
\item \texttt{thr\_fixed}: F1 gap under the fixed-threshold variant
  (P3M lacks this contract by construction).
\end{itemize}
The held-out contracts are \texttt{gold\_gR} (F1 gap under the
Re-DocRED revision $g_R$) and \texttt{thr\_cal}: the F1 gap computed
under per-class sigmoid thresholds re-fitted on a held-out
calibration split against the $g_R$ golds (the harness's
\texttt{calib} objective, gold-bearing-only tie rule)---the
threshold and gold axes therefore share calibration data by
construction, and the decode-calibration probe (prior-matched,
gold-free thresholds) exists precisely to audit this interaction.
The axis generalizes ATLOP's adaptive thresholding
\cite{zhou-etal-2021-atlop} to a declared contract coordinate. All
contracts share the frozen earliest-mention grouping rule and the
frozen per-document counts.

\bibliographystyle{sn-basic-cite}
\clearpage
\bibliography{followup_refs}

\end{document}